\documentclass[12pt]{article}

\usepackage{latexsym}
\usepackage{graphicx}
\usepackage{multicol,multirow}
\usepackage{booktabs}
\usepackage{amsmath,amssymb,amsfonts}
\usepackage{mathrsfs}
\usepackage{amsthm}
\usepackage[numbers,sort&compress]{natbib}
\usepackage{rotating}
\usepackage{appendix}
\usepackage{ifpdf}
\usepackage[T1]{fontenc}
\usepackage{times}
\usepackage{textcomp}%
\usepackage{xcolor}%
\usepackage{hyperref}
\usepackage{float}
\usepackage{subfig}
\usepackage{IEEEtrantools}
\usepackage[a4paper,margin=1in]{geometry}

\newenvironment{fullwidthstatement}[1]
{\par\noindent\begin{center}\textbf{#1}\end{center}\noindent}
{\par}

\title{From Grey-Box to Green-Box: When can Physics-Informed
Machine Learning Reduce Carbon Footprints in Structural
Health Monitoring?}

\author{Daisy R Bradley\textsuperscript{1,*} \quad Matthew R Jones\textsuperscript{1} \quad Nathan A Hinchliffe\textsuperscript{1}\\
Daniel J Pitchforth\textsuperscript{1} \quad Elizabeth J Cross\textsuperscript{1}\\[0.5em]
\small\textsuperscript{1} School of Mechanical, Aerospace and Civil Engineering,\\
\small University of Sheffield, UK\\
\small\textsuperscript{*}Corresponding author: \href{mailto:dbradley2@sheffield.ac.uk}{dbradley2@sheffield.ac.uk}}
\date{}

\begin{document}

\bstctlcite{IEEEexample:BSTcontrol}

\maketitle

\begin{fullwidthstatement}{Abstract}
As machine learning becomes central to modern engineering, the significant compute requirements of black-box models impose a substantial, often overlooked, environmental cost. This study provides the first exploration of physics-informed machine learning (PIML), or “grey-box” modelling, as a pathway towards sustainable machine learning for engineering, via reduced training data volume. In a structural health monitoring context, we evaluate four key PIML techniques - residual modelling, input-augmentation, hybrid modelling, and constrained learning - against conventional black-box baselines, considering the training data required to achieve a given accuracy threshold. Beyond training, we qualitatively consider how PIML compounds carbon savings by diminishing data collection and storage infrastructure requirements. Grey-box models are shown to frequently require less training data to reach or surpass this performance target, decreasing training time and therefore carbon emissions as a result.
\end{fullwidthstatement}

\vspace{\baselineskip}

\begin{fullwidthstatement}{Impact Statement}
The use of artificial intelligence and machine learning within the engineering sector is skyrocketing, alongside the environmental impact associated with engineering computing. Physics-informed machine learning combines data-driven methods with physical insight of the system in question, often leading to increased extrapolation performance and therefore reduced training data requirements. This work considers physics-informed machine learning as a potential route to decrease modelling emissions based on a training data reduction compared to a black-box equivalent model, as data collection, storage, and use in training can be minimised, while maintaining or even improving model performance.
\end{fullwidthstatement}

\section{Introduction}
\label{sec:introduction}

Within the engineering community, our use of machine learning (ML) and artificial intelligence (AI) is significantly increasing, leading to a corresponding rise in the energy demand from our computing.
This comes with wide ranging and well documented environmental impacts, including increased carbon emissions.
So far, however, there has been little consideration of how this will accelerate further as we increase our use of these tools in a professional context, particularly where we might expect considerable interplay between AI and other approaches with a heavy computational burden, such as computation fluid dynamics (CDF) or finite element analysis (FEA).
With the main AI providers not currently required to share their energy and water use, barriers to quantifying and projecting this impact are large.
Regardless, there are steps we can take as engineers to mitigate the 
potential impact.

Within the broad remit of our AI use, this work considers the impact of 
machine learning models used for inference in an engineering or scientific 
context (e.g. for prediction of key 
structural response variables). Following a growing interest in 
physics-informed machine learning, and noticing its benefits in particular in 
reducing the burden on training data \cite{cross2024spectrum}, we introduce 
and ask the question here as to whether embedding 
physics directly into a machine learning model could help to reduce its 
environmental impact. In this paper we will explore when this may or may 
not hold true with examples from regression challenges in structural health 
monitoring (SHM) (an area where physics-informed models have been shown to 
outperform their black-box counterparts).

Our particular focus is on the relationship between training data quantity 
and training time, linking this duration to carbon emissions from computing, 
and exploring the hypothesis that reducing training data requirements reduces 
the environmental impact of a model via shorter training duration.
We also consider the carbon cost of collecting and storing additional 
training data.

Physics-Informed Machine Learning (PIML) (also known by several variant 
names \cite{cicirello2024physics,haywood2024discussing}) offers different 
ways to embed physical knowledge 
into a model\footnote{Note that we will often refer to models which are 
purely data-driven as black-box, those that are purely physical as white-box, 
and those in-between as grey-box}. 
PIML techniques exist on a spectrum, where "whiter" models contain more physical insight, and "blacker" models are more data-driven.
To evaluate when this improves model 
sustainability, we explore a range of PIML approaches with varying degrees of 
embedded physics. We will employ neural networks (NNs), but predominantly Gaussian 
Processes (GPs) as regression methods in our comparisons. Although GPs are 
known to be slow 
because of their $\mathcal{O}(N^3)$ computational overhead (without sparse 
approximation methods), they offer a means to embed different levels of 
physical insight within the same mathematical framework. This provides a 
baseline model for comparison, allowing us to evaluate how incorporating 
physics impacts model cost. Explicitly, we will consider \textit{residual 
modelling}, where the black-box component accounts for un-modelled behaviour 
in the physics component, \textit{input augmentation}, where we adapt input 
variables to account for known physics before feeding them into a black-box 
regressor, \textit{hybrid models}, where the explanatory power 
is shared more equally between the physics and data components (here through 
GP kernel design), and finally \textit{constrained learners}, where the 
black-box 
learner is dominant in the model but restricted by physical constraints.

Throughout, a common theme will be the consideration that in many engineering 
applications, we have some idea of the required level of performance - how 
good is good enough? This work is the first to draw comparisons between 
``grey-box'' and black-box models through an environmental lens in the 
pursuit of ``green-box'' models which meet the required performance with 
minimal environmental impact.

\section{Context and literature}
\label{sec:lit_review}

The world has now entered into the ``Age of Artificial Intelligence'', where AI has gone from an abstract concept to something that many of us use on a daily basis.
Within both industry and academia, AI tools are being heartily adopted and deployed, and while this has brought about astonishing advancements across many sectors \cite{sajadieh2026aiindex}, the detrimental impact to the environment cannot be ignored.
Data centres used to power AI cause many environmental concerns 
\cite{selvan2025sustainable}, ranging from the toxic runoff from the 
extraction of rare earth elements \cite{giglio2021extractivism} to 
manufacture graphics processing units (GPUs), the embodied carbon within the 
hardware itself \cite{luccioni2023estimating}, and the significant amounts of 
water lost for evaporative cooling of data centres in water-stressed areas 
\cite{li2023making}.
The energy used to power these data centres is already significant, estimated 
at 415 TWh in 2024, and predictions from the International Energy Association 
suggest that this could rise to 945 TWh by 2030 - which is comparable to the 
annual electricity consumption of the continent of Africa 
\cite{selvan2025sustainable} - creating a predicted 320 Mt of 
carbon dioxide \cite{iea2025energyai}. 

Machine learning methods are a subset of AI which learn patterns from 
data, and are commonly used within engineering and computer science. Concerns 
around the impact of AI have given rise to a new field of research 
focusing on sustainable ML, where it is well recognised that accurate 
measurement of computational carbon footprints is a prerequisite for 
emissions reduction \cite{lannelongue_ten_2021}. Numerous software frameworks 
have been developed to quantify computing energy usage which can be divided 
into those which use static estimation techniques that rely on fixed baseline 
assumptions, and dynamic estimation methods that capture real-time usage data 
from central processing units (CPUs) and GPUs \cite{fischer2025groundtruth}. Prominent static tools 
include the Machine Learning Emissions Calculator (MLCO2) 
\cite{lacoste2019quantifying}, whereas utilities such as CodeCarbon 
\cite{codecarbon_methodology} and eco2AI \cite{Budennyy2022eco2AI} 
represent dynamic tracking implementations. Although these tools are 
extensively integrated into green computing research, comparative benchmark 
studies evaluating them against physical power meters reveal notable variance 
and a systematic tendency to under-report emissions \cite{Bouza_2023, 
fischer2025groundtruth}. Nevertheless, applying a chosen tracking methodology 
consistently provides a valid baseline for relative comparison, guiding more 
environmentally conscious model design and deployment decisions.

In terms of actually reducing the footprint of ML, there is extensive work in 
the literature on increasing the efficiency of ML models, although not 
generally with a view to reducing emissions. In \cite{Menghani_review_2023}, the 
author categorises efficient techniques into five focus areas: compression 
techniques, learning techniques, automation, efficient architectures, and 
infrastructure and hardware.
Some of the most successful techniques include pruning, quantisation, and mixture of expert (MoE) architectures. 
Pruning is a compression technique which has been successful at downsizing ML models without compromising performance; in \cite{han2015learningweightsconnectionsefficient}, the total number of parameters was reduced by a factor of 13 with no loss of accuracy.
Quantisation is another popular compression technique which has successfully reduced both inference time and storage requirements by reducing the precision used in neural network model \cite{Gholami_Kim_Dong_Yao_Mahoney_Keutzer_2021}.
In several instances, the quantised model shows negligible accuracy loss compared to the uncompressed model \cite{sharify2024posttrainingquantizationlarge}.
The efficient MoE architecture has been widely adopted, with one of the most famous examples being the DeepSeek-V3 model \cite{deepseekai2025deepseekv3technicalreport}. 
This was built on the DeepSeekMoE architecture, which achieves comparable performance to Meta's LLaMA2 7B with only 40\% of its computations \cite{dai2024deepseekmoeultimateexpertspecialization}.
These developments showcase a small subset of the ongoing work in the field of efficient ML.  
However, Jevons' paradox states that as technology makes resource use more efficient, total consumption of that resource increases rather than decreases \cite{jevons1865coal}.
To avoid this being the case for ML, it is imperative that we keep the broader context of sustainability at the forefront of research and development.

Again, although not necessarily for environmental reasons, considerable 
research has 
focused on enhancing the computational efficiency of traditional numerical 
models in engineering settings. Reduced order methods (ROMs) significantly 
lessen the computational burden in time-critical applications such as 
real-time control, structural dynamics, and digital twins 
\cite{quarteroni2014reduced, benner2015survey}. Alternatively, surrogate 
modelling using statistical methods (e.g. GPs) offers a data-driven 
approach to approximate high-fidelity, computationally expensive simulations 
\cite{sacks1989design, kudela2022surrogate}.
 
As engineers, we usually work in environments where we have both data (real or simulated) and a physical understanding of the system we are modelling, which can be combined.
As we move towards incorporating physics knowledge into ML, we have the potential to increase our sustainability even more by combining both efficient ML techniques and efficient numerical techniques to reduce our computational demand.

PIML is the subcategory of ML where physical knowledge is integrated into the 
models to form so-called ``grey-box'' models. 
Some of the benefits of grey-box models compared to their black-box counterparts include increased generalisation, efficiency and interpretability \cite{Meng2025WhenPhysicsMeetsML}.
In an SHM context, grey-box models have been shown to outperform black-box models in situations where training data does not span all environmental and operational conditions (EOCs) \cite{cross2022physics}.
Here, the relationship between physics input and training data requirements 
will be further explored and directly linked to environmental impact.

Of the numerous available methods for PIML \cite{Meng2025WhenPhysicsMeetsML}, 
physics-informed neural networks (PINNs) remain the most widely adopted, 
integrating physical principles by incorporating the residual from known 
differential equations directly into the loss function. Against our 
hypothesis, existing literature demonstrates that introducing physics into 
the loss of a neural network increases computational burden. For example, in 
\cite{stiasny2023physics}, black-box NNs are compared to 
PINNs to accelerate power system time-domain dynamic simulations.
While both ML methods were shown to be faster than a traditional numerical 
solver, the up-front training cost per epoch were higher for the PINN due to 
evaluating the physics loss term.
\cite{fernandes2026computational} also find the computational cost of PINNs 
to be higher than traditional NN solvers in the context of immune system 
response modelling, due to the
repeated evaluation of differential operators. Compounding these raw 
computational costs, PINNs 
frequently suffer gradient ``pathologies" (e.g. vanishing gradients) during 
optimisation 
\cite{wang2020understanding}, leading to slow convergence and further 
inflating total training time and energy consumption. Noting these results, 
and not wishing to add to the further proliferation of PINNs, this paper will 
focus on
other PIML methods which have not yet been explored in terms of their 
computational cost. 

\subsection{Our Approach}

The hypothesis we are interested in exploring is whether embedding physics into 
a machine learner can reduce its emissions. We have already read how 
physics in the loss of a PINN can certainly do the opposite. Here we more 
broadly consider a range of PIML methods, where physical insight/models have 
varying strengths in the inference.

Many of the examples are drawn from the 
work of the authors and rely heavily on a Gaussian process framework to 
introduce physical insight in different ways. The models are developed for 
the purposes of SHM, where the main gain that we 
have seen from PIML is a reduction in the need for training data covering the 
whole operational span of a structure, or from all different environmental 
conditions \cite{cross2022physics}. In this work we will explore how the 
gains you can make from 
reducing your data requirements (which here we refer to as \textit{coverage}) 
do or do 
not also reduce emissions when training the model. For each test case 
presented we will show how the model error changes with coverage using learning curves \cite{viering2023shape} and 
how the runtime of that model is affected. We will explore our hypothesis by 
investigating five PIML 
techniques - ordered in terms how dominant the physics component is - from 
most to least: 

\begin{itemize}
	\item Residual modelling: a physical model is employed, with 
	a machine 
	learner accounting for the error between the prediction of the physical 
	model and the target. In Section~\ref{sec:residual}, we use a zero mean Gaussian 
	process with an 
	Squared Exponential (SE) kernel, and in comparison, a multi-layer perceptron.
	\item Input augmentation: using physical insight to transform inputs before they are passed to a machine learner. Transformed inputs may replace or be used alongside existing measured data, with the aim of simplifying a learning task. In Section~\ref{sec:input_augmentation} we apply the same GPs and NNs 
	as above, with 
	and without the `manufactured' inputs.
	\item Hybrid modelling through the GP kernel: Here we 
	bring our knowledge 
	in through adapting the kernel/covariance function that dictates how our 
	functions behave. In Sections~\ref{sec:conference} and~\ref{sec:Dan}, we consider two kernel 
	designs, one that 
	can switch between different physical states.
	\item Constrained GPs: At the blackest end of the grey-box 
	spectrum, in Section~\ref{sec:Matty}
	we take advantage of a sparse approximation that allows us to embed the 
	known boundary conditions of the problem into the covariance function.
\end{itemize}

\noindent \textbf{Caveats:} In this 
initial work, we are investigating the examples where we have seen performance 
gain by introducing physics.  The implicit assumption we have made 
is that if physics does not prove to be useful in terms of performance, then 
it will not help reduce emissions. The second caveat here is that for this 
initial work, we are exploring our hypothesis with what would be considered 
very small datasets in a deep learning context - this will be discussed 
further in later sections.

In the remainder of the paper, Section~\ref{sec:methods} will introduce the specific models we 
will consider, along with implementation details. The results of each of the 
methods considered are presented in Sections ~\ref{sec:residual} through ~\ref{sec:Matty}, while Section~\ref{sec:conclusions} concludes 
the paper and outlines directions for future work. As we go through the case 
studies, we will explore our initial expectation that increasing the 
contribution of the physical component in the machine learner should in turn 
decrease the computational burden.

\section{Methods}
\label{sec:methods}

\subsection{Models}
\label{sec:models}

In this work, both GPs and NNs are 
used as black-box models. Readers are referred to \cite{RasmussenWilliams2006} 
for a detailed explanation of GP methods, and to 
\cite{gurney1997introduction} for NNs. The primary model here is a GP, as it 
allows us to easily embed varying levels of physical insight to explore our 
hypothesis.

A GP is a non-parametric statistical model, whereby predictive inference is achieved from a Bayesian update of the prior mean and covariance with the training data (X, Y), and evaluated at test points (X*) to obtain predictions (Y*). The update equations are shown in Equations \ref{eq:GP_e} 
and \ref{eq:GP_v}.

\begin{equation}
	\mathbb{E}[f*]=K(X_*,X)[K(X,X)+\sigma_n^2I]^{-1}y
	\label{eq:GP_e}
\end{equation}

\begin{equation}
	\mathbb{V}[f_*]=K(X_*,X_*)-K(X_*,X)[K(X,X)+\sigma_n^2I]^{-1}K(X,X_*)
	\label{eq:GP_v}
\end{equation}

\noindent where $K(X,X)$ is the covariance matrix for the training 
inputs, $K(X_*,X)$ the covariance matrix of the testing and training inputs, 
and $K(X_*,X_*)$ the covariance of the testing inputs, and $\sigma_n$ is a 
noise hyperparameter.

The GP prior is comprised of two components, the mean and the covariance, defined by a covariance/kernel function.
Both the mean and covariance functions can be used to embed physical insight into the model.
Residual modelling, for example, incorporates a non-zero 
prior mean function.
Many different kernel functions exist to impose different behaviours, such as smoothness or periodicity, via control of their hyperparameters (see \cite{duvenaud2014automatic}).
These hyperparameters are learned through optimisation of the negative 
marginal log likelihood - optimisers used in this work include gradient descent algorithms and quantum particle swarm optimisation (QPSO), depending on the case study (as will be highlighted).
Standard black-box modelling typically combines 
a zero mean with the SE kernel shown in 
Equation \ref{eq:SE_kernel}.

\begin{equation}
	k_{SE}(x,x') = \sigma^2exp(-\frac{(x-x')^2}{2l^2})
	\label{eq:SE_kernel}
\end{equation}

\noindent where $\sigma^2$ is the output variance hyperparameter and $l$ is the lengthscale hyperparameter.

Kernel selection can be guided by known system behaviours - an approach termed physics-informed kernel selection. 
Furthermore, hyperparameters can be bounded or fixed to reflect these 
behaviours, reducing the search space to optimise. Kernels can also be 
combined to embed domain knowledge, such as the switching kernels from 
\cite{pitchforth2026physics} explored in Section~\ref{sec:Dan}. GPs can also be 
constrained to comply with physical laws or assumptions — for 
instance, enforcing non-negativity through the GP prior 
\cite{pensoneault2020nonnegativity} or monotonicity using virtual derivative 
observations with a probit likelihood function \cite{riihimaki2010gaussian}. 
Constrained GPs in acoustic emission mapping \cite{jones2023constraining} 
serve as a case study in Section~\ref{sec:Matty}.

The NNs used in this work are multilayer perceptrons (MLP) with 3 
hidden layers, 64 neurons per layer, and three ReLu activation functions.
Optimisation uses ADAM \cite{kingma2017adammethodstochasticoptimization} with 
a maximum training budget of 1000 epochs and an early stopping triggered if 
the validation performance fails to improve by at least $10^{-5}$ for 20 
consecutive epochs.

For all models in this paper, a normalised mean squared error (NMSE) is used 
to measure performance, as defined by Equation \ref{eq:NMSE}

\begin{eqnarray}
	NMSE = \frac{100\Sigma(y-y*)^2}{n\sigma}
	\label{eq:NMSE}
\end{eqnarray}

\noindent where $y$ represents the true values, $y*$ the predicted values, 
$n$ is the 
number of data points in the testing set and $\sigma$ the variance of the 
true values.
Throughout, we will take consideration that, as engineers, we may have an 
idea of the level of performance the 
model must achieve to be useful in a particular context, in some examples we 
will use this as a threshold through which we can then compare performance. 

\subsection{Model Carbon Emissions}

The methodology used to approximate the carbon emissions of the models 
closely follows the methodology used by CodeCarbon 
\cite{codecarbon_methodology}.
In each test case, the model is run purely on a CPU, with the CPU approximated to be operating at 50\% of its thermal 
design power (TDP).
The power associated with random access memory (RAM) is based on the number 
of modules in the workstation, with 32 GB of RAM (the capacity of all 
workstations used in this study) estimated to use 20 W 
\cite{codecarbon_methodology}.
Equation~\ref{eq:energy} shows how these constant power estimations lead to 
energy scaling linearly with time.

\begin{equation}
	E_{total} = t (0.5P_{TDP}+ 20)
	\label{eq:energy}
\end{equation}

\noindent Where $E_{total}$ is the total energy consumed in kWh, $t$ is the time taken to run the model in hours, and $P_{TDP}$ is the thermal design power of the CPU in kilowatts.
As all models were run in the UK, a carbon intensity value of 
237.589 
$gCO_{2}eq/kWh$ was used to convert from energy into emissions 
\cite{codecarbon_global_energy_mix}. 
Due to the linear relationship between time and emissions, discussions will 
be framed in terms of run time.

An additional consideration is the carbon cost of collecting and storing 
training data.
Data collection ranges from little to significant carbon cost, depending on 
the the problem. For example, while pre-existing bridge sensors carry an 
embedded carbon cost from their manufacturing and installation, collecting 
data from them incurs virtually no additional operational emissions. In 
contrast, gathering aircraft data may require dedicated flights, generating 
operational emissions that easily dwarf those from training the model itself.
While it is impossible to quantify the carbon cost of data collection in a 
generic case, it is something that we must keep in mind when considering our 
ML modelling.
The carbon cost of data storage is also difficult to quantify, as it depends 
heavily on whether storage is local or cloud-based, and if it is stored in a 
data centre, the efficiency and energy source of the data centre.
\cite{adamson2017carbon} estimates that transmitting and storing data in a 
data centre requires 3-7 kWh per GB, and that storing 100 GB of data for a 
year costs approximately 0.2 tons of $CO_{2}$.
Though this estimation is very loose, it highlights that the carbon cost of 
data storage can be significant, especially for the vast datasets used in 
SHM, as explored in \cite{brennan2025foundations}.

The next section of this paper will consider residual modelling as our first 
PIML example, where we will compare a residual model against a black-box 
model for both NNs and GPs, and explore the relationship between physics 
input, training data, performance, and emissions.

\section{Residual Modelling}
\label{sec:residual}

Residual models are a PIML method at the whiter end of the spectrum with 
physics playing the most dominant role, where a physics-based or ``white-box'' 
model is first implemented to capture the known relationships within the data.
The white-box model output is then subtracted from the observed data to give a residual, which is then used as a target for a data driven model.
In engineering, we often understand some underlying physics of our system, 
making residual modelling a good candidate for many applications.

The models used in this case study were developed based on the work in 
\cite{zhang2021gaussian} to predict longitudinal deck deflections of the 
Tamar bridge, as this can be used as a performance indicator 
\cite{cross2012structural}. The changes in deck deflection are driven by a 
number of factors, predominately temperature, wind conditions and traffic 
load.
The available data spans from September to January, therefore capturing both 
daily fluctuations and the seasonal temperature change.
As it is known that stay cables expand linearly with temperature 
\cite{westgate2012environmental}, a linear model was fit to the first 500 
temperature and deflection data points to serve as the physics component. For 
the grey-box models, a GP or a NN was then used to learn the 
residual between the prediction of the physics component and the measured 
deflection. The black-box models simply learned the relationships between 
input and output data directly; here, the GP has been restricted to a 
temperature only input, while the NN has inputs of temperature, 
traffic loading and wind speed. The GP input is restricted because the 
black-box GP model showed poor performance when learning with all three 
inputs, hence why temperature was used alone, allowing for a fair comparison 
with the grey-box GP.
Both models use ADAM optimisation \cite{kingma2017adammethodstochasticoptimization} to learn the hyperparameters.

This model was established to explore how useful the grey-box approach is in 
the case of limited data available for model training. It is often the case that data 
for a monitoring system are available only for a fixed time, and will not 
cover all operational/environmental conditions that a structure will see in 
its life time. Here we explore the performance of the models with different 
training periods - which in this case relates to different coverage of the 
environmental conditions. The data reflect a time when the weather is getting 
progressively colder, so models trained on only, say the first 10\% of the 
data, are 
required to extrapolate into the colder temperatures - this is where the 
linear physics increases the grey-box performance. In terms of the impact of 
establishing the models, we expect that the ability to work with less 
training data should lead to reduced training time and therefore emissions, 
as well as those from the collection and storage of data.

For the results here, the data were split into 10\% coverage increments 
temporally. Training data were increased by adding 10\% increments until full 
coverage was reached. Within the training coverage area, the GPs used a 50/50 
train/test split the NNs used a 50/15/35 train/validation/test split.
The training times reported for this study do not include the time to train 
the linear model, as it is only run once as a pre-processing step prior to 
model training, and was not repeated for each coverage increment.

\subsection {Results and Discussion}

\begin{figure}[h]
    \centering
    \subfloat[NMSE]{%
        \includegraphics[width=0.48\linewidth]{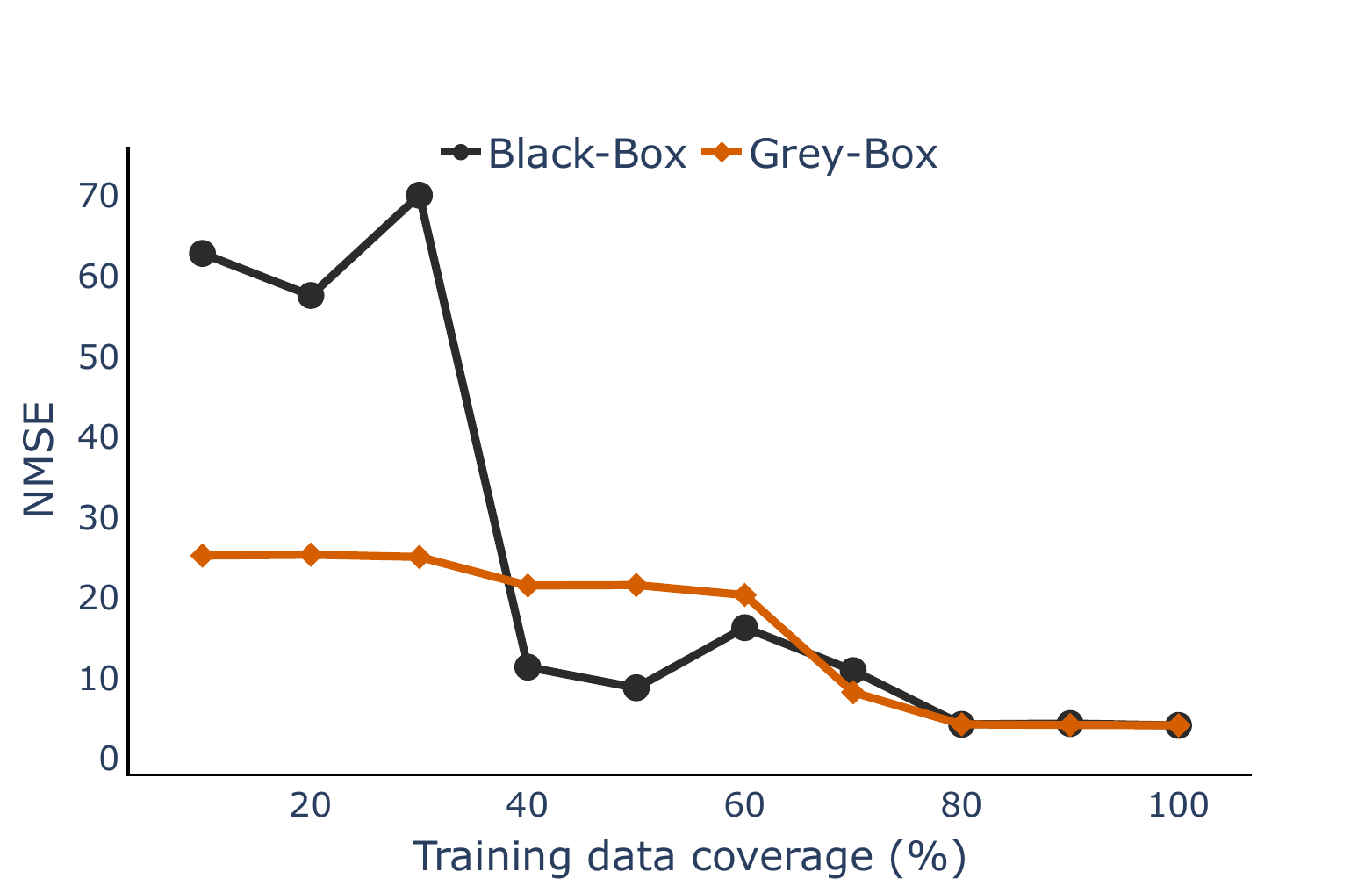}%
        \label{fig:daisy_gp_tamar_nmse}%
    }
    \hfill
    \subfloat[Training Time]{%
        \includegraphics[width=0.48\linewidth]{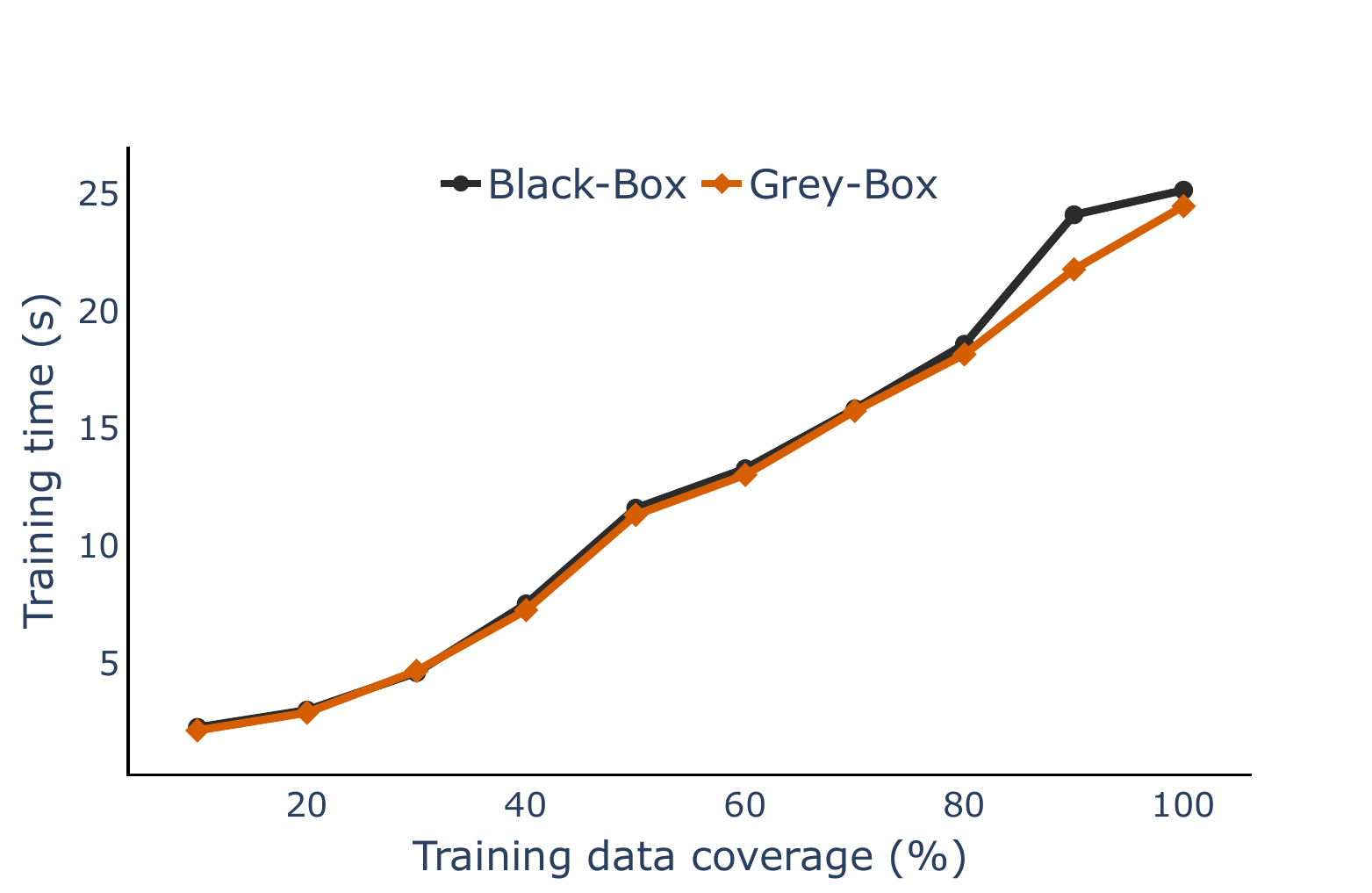}%
        \label{fig:daisy_gp_tamar_time}%
    }
    \caption{The NMSE and training time results for the GP case study, showing the performance of the black-box and residual models.}
    \label{fig:Daisy_GP_Tamar_results}
\end{figure}

\begin{figure}[h]
    \centering
    \subfloat[NMSE]{%
        \includegraphics[width=0.48\linewidth]{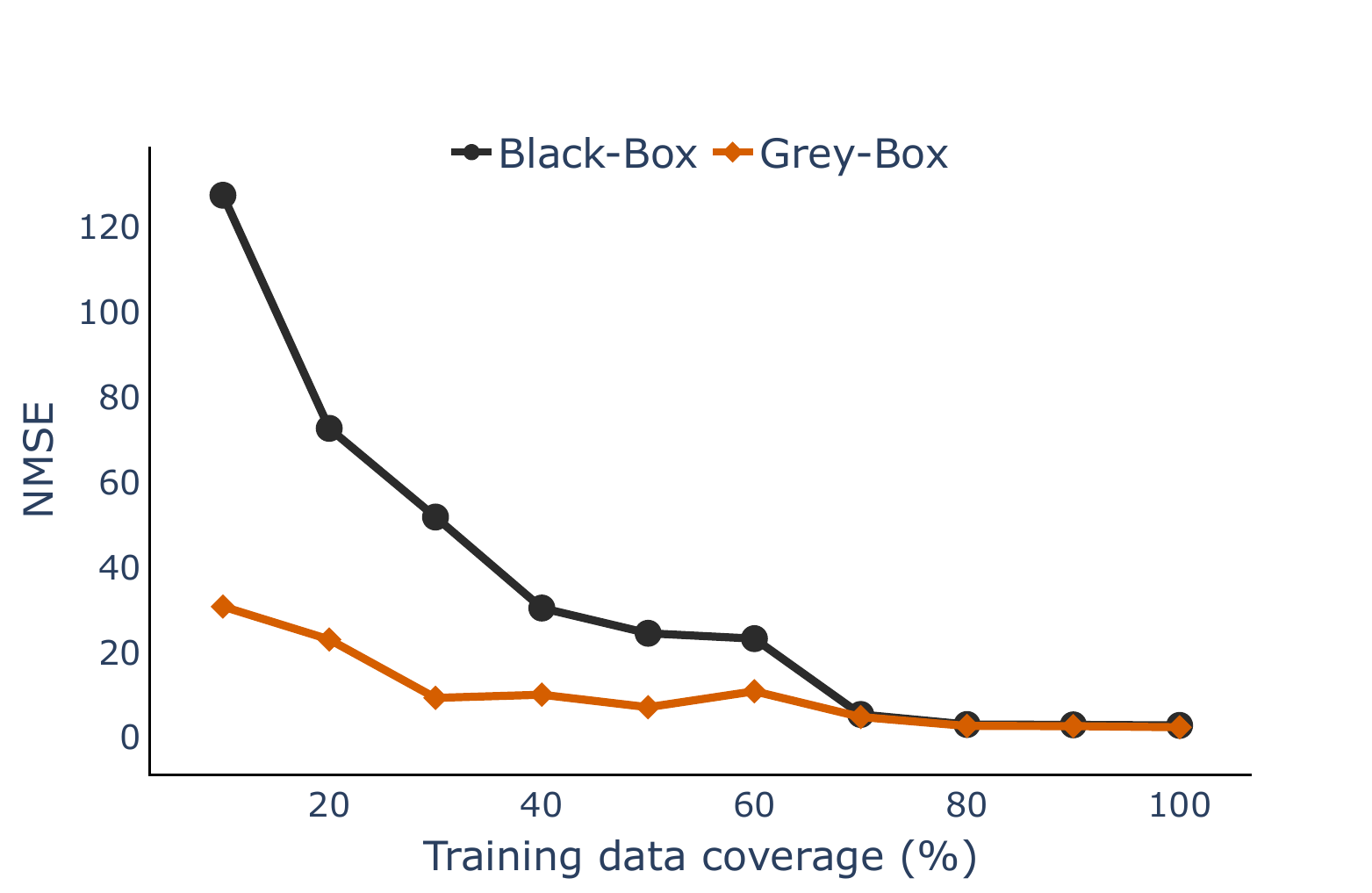}%
        \label{fig:daisy_nn_tamar_nmse}%
    }
    \hfill
    \subfloat[Training Time]{%
        \includegraphics[width=0.48\linewidth]{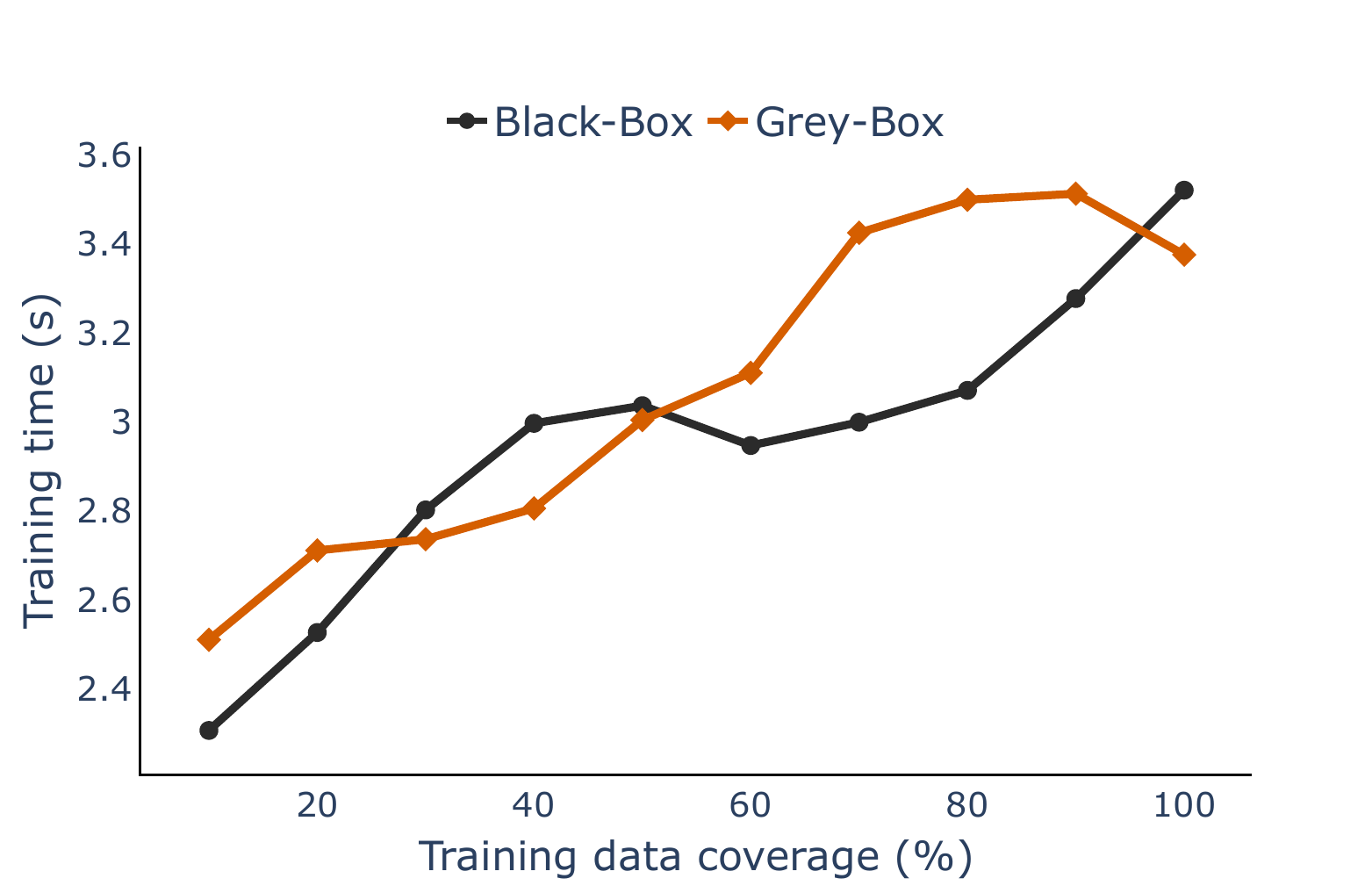}%
        \label{fig:daisy_nn_tamar_time}%
    }
    \caption{The NMSE and training time results for the NN case study, showing the performance of the black-box and residual models.}
    \label{fig:Daisy_NN_Tamar_results}
\end{figure}

Figure \ref{fig:Daisy_GP_Tamar_results} shows the results for the black-box 
and grey-box GP models, comparing them in terms of NMSE and training time, 
against increasing coverage of training data, while Figure 
\ref{fig:Daisy_NN_Tamar_results} shows the same for the NN models.

\begin{figure}[h]
    \centering
    \includegraphics[width=0.6\linewidth]{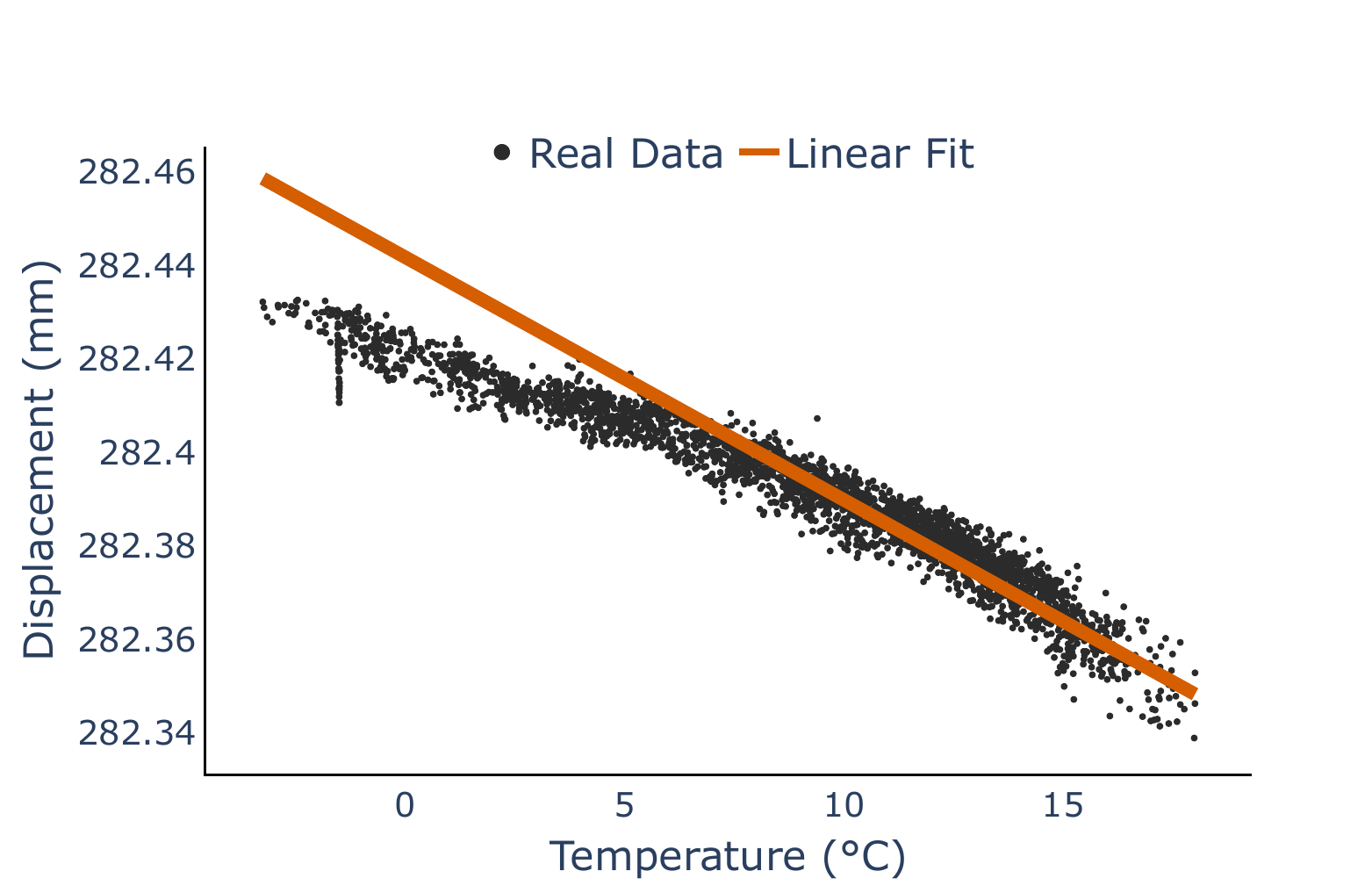}
    \caption{The linear model fit to the Tamar bridge deflection data, showing that the linear model does not fully capture the relationship between temperature and deflection.}
    \label{fig:tamar_linear_model}
\end{figure}

As the black-box models cannot extrapolate, they were expected to perform poorly at low training data coverage, where training data is from September/October, excluding the cold temperatures seen in December/January, as can be seen in Figures~\ref{fig:daisy_gp_tamar_nmse} and \ref{fig:daisy_nn_tamar_nmse}.
The grey-box models will outperform the black-box models in these training 
data regimes, as the linear physics allows for extrapolation to cooler 
temperatures, although, as expected, we see a performance levelling between 
the two models 
as training data coverage increases and cooler temperatures are included in 
the training data.
While the NN models follow this expected behaviour, as shown in 
Figure~\ref{fig:daisy_nn_tamar_nmse}, Figure~\ref{fig:daisy_gp_tamar_nmse} 
shows the black-box outperforming the grey-box at medium coverage (40-60\%). 
This suggests that the physics-based linear model does not totally encompass 
the relationship between temperature and deflection, which is supported by 
Figure~\ref{fig:tamar_linear_model}, which plots the displacement of the deck 
against temperature and shows the linear model fit to the data.  Another 
consideration is that we have removed the simple-to-learn linear 
relationship from the observed data, meaning that the residual model may 
have a more complex relationship to learn, such that in this medium coverage 
range, the grey-box GP struggles to perform.

When comparing the run times of the models (and noting that they are small 
due relatively small dataset), we can see the expected behaviour of these 
increasing with training data size. Loosely we would expect the GP to scale 
cubically with training data size, and the neural network to scale linearly.

In terms of making a comparison between them in terms of emissions, we 
consider different NMSE \textit{thresholds} - to compare the emissions 
between the models when they are required to hit a pre-determined level of 
fidelity.

Unlike the NN case, where the grey-box model would reach \textit{any} NMSE 
threshold set with less training data than the black-box, our NMSE threshold 
affects which model is more effective in the GP case.
If we take a threshold of NMSE $<$10, the grey-box GP model reaches this with 70\% coverage, as opposed to 30\% coverage for the black-box, which leads to a 38.7\% increase in carbon emissions from training (0.31 gCO$_2$e for the black-box vs 0.43 gCO$_2$e for the grey-box).
However, a threshold of NMSE $<$30 sees the grey-box model reach this with only 10\% coverage, giving a 50.0\% reduction in carbon emissions compared to the black-box, which requires 30\% coverage to reach the same performance (0.06 gCO$_2$e for the grey-box vs 0.12 gCO$_2$e for the black-box).
This demonstrates the importance of deciding what performance is ``good enough'' as early as possible, making sure we only collect, store and use as much data as we need, and allowing us to decide the most appropriate model for our needs.

Considering Figure~\ref{fig:daisy_gp_tamar_time}, we can see that the training time for the two GP models is nearly identical, as expected considering the models have the same hyperparameter count and a similar problem complexity.
In some cases, there is a trade off between hyperparameter count and training data requirements when introducing physics (see Sections~\ref{sec:conference} and \ref{sec:Dan} for examples) however, in this study there is no run time penalty from adding physics, so any training data reduction will reduce training emissions.
Similar closely matched training times can be seen in Figure~\ref{fig:daisy_nn_tamar_time}, though in this case, the times are fast enough that carbon savings from model training would be minimal.
For example, an NMSE threshold of $<$30 sees a 10.7\% reduction in emissions, even though the grey-box model needs less than half of the training coverage of the black-box, which equates to a saving of 0.009 gCO$_2$e.
Nonetheless, in both cases, collecting and storing less data will reduce the overall carbon impact of the models.

This case study has shown that residual models can outperform black-box models, especially in regions of low training data coverage, which can lead to reduced training time and therefore emissions from training, as well as those from the collection and storage of data, when less training data is required to reach a given performance threshold.
While these results are promising, it is important to note that the performance of a residual model is highly dependent on the ability of a physics-based model to capture underlying relationships in the data, though there are many cases where our engineering knowledge makes this possible.

In the following section, we will explore input augmentation, an approach with a less predictable outcome.

\section{Input Augmentation}
\label{sec:input_augmentation}

Input augmentation involves preprocessing our data, often translating from 
raw sensor data into more meaningful physical quantities, before using these 
as the inputs to a data-driven model.
These physics-based augmentations can simplify the ML problem if the new 
inputs have simpler relationships to the target, potentially reducing the run 
times requirements for ``good'' model performance, leading to carbon savings 
for the model.

This input augmentation case study was inspired by the work of \cite{reed2005parametric,gibson2023distributions}, in which the authors modify sensor data using known physical relationships of an in-service aircraft and use these as model inputs to predict wing strain, using a NN and a GP respectively.
In this study, a black-box and grey-box version of both a NN and GP are constructed and compared, with architectures described in Section~\ref{sec:models}, and with the inputs differentiating the black-box and grey-box models.
The GP models use a  limited-memory Broyden–Fletcher–Goldfarb–Shanno (L-BFGS) optimiser to learn the hyperparameters, while the NN models use ADAM optimisation \cite{kingma2017adammethodstochasticoptimization}.
To provide input for the black-box models, the raw data from 7 sensor channels were used.
These same 7 channels were augmented to create inputs for the grey-box 
models, which includes using Newton's second law of motion to convert 
accelerations into forces with total aircraft mass, dynamic pressure 
calculated from the indicated airspeed and the air density (estimated from the 
barostatic altitude) as in \cite{anderson2016introduction}, and the 
directional accelerations calculated based on aircraft specific formulae.
The complete set of inputs for each model are shown in Table~\ref{tab:nathan_inputs}.

\begin{table}[h]
    \centering
    \caption{Black-box and Grey-box model inputs for the prediction of the 
    strain measured on an aircraft wing during flight}
    \label{tab:nathan_inputs}
    \begin{tabular}{ll}
        \toprule
        Black-Box Inputs & Grey-Box Inputs \\
        \midrule
        Normal acceleration at the CG & Normal force at the CG \\
        Lateral acceleration at the CG & Lateral force at the CG \\
        Normal acceleration at the empennage & Pitch acceleration \\
        Normal acceleration at the port wing & Roll acceleration \\
        Indicated airspeed & Dynamic pressure \\
        Barostatic altitude &   \\
        Remaining fuel mass &  \\
        \bottomrule
    \end{tabular}
\end{table}

\subsection{Results and Discussion}

\begin{figure}[h]

    \centering
    \subfloat[NMSE]{%
        \includegraphics[width=0.48\linewidth]{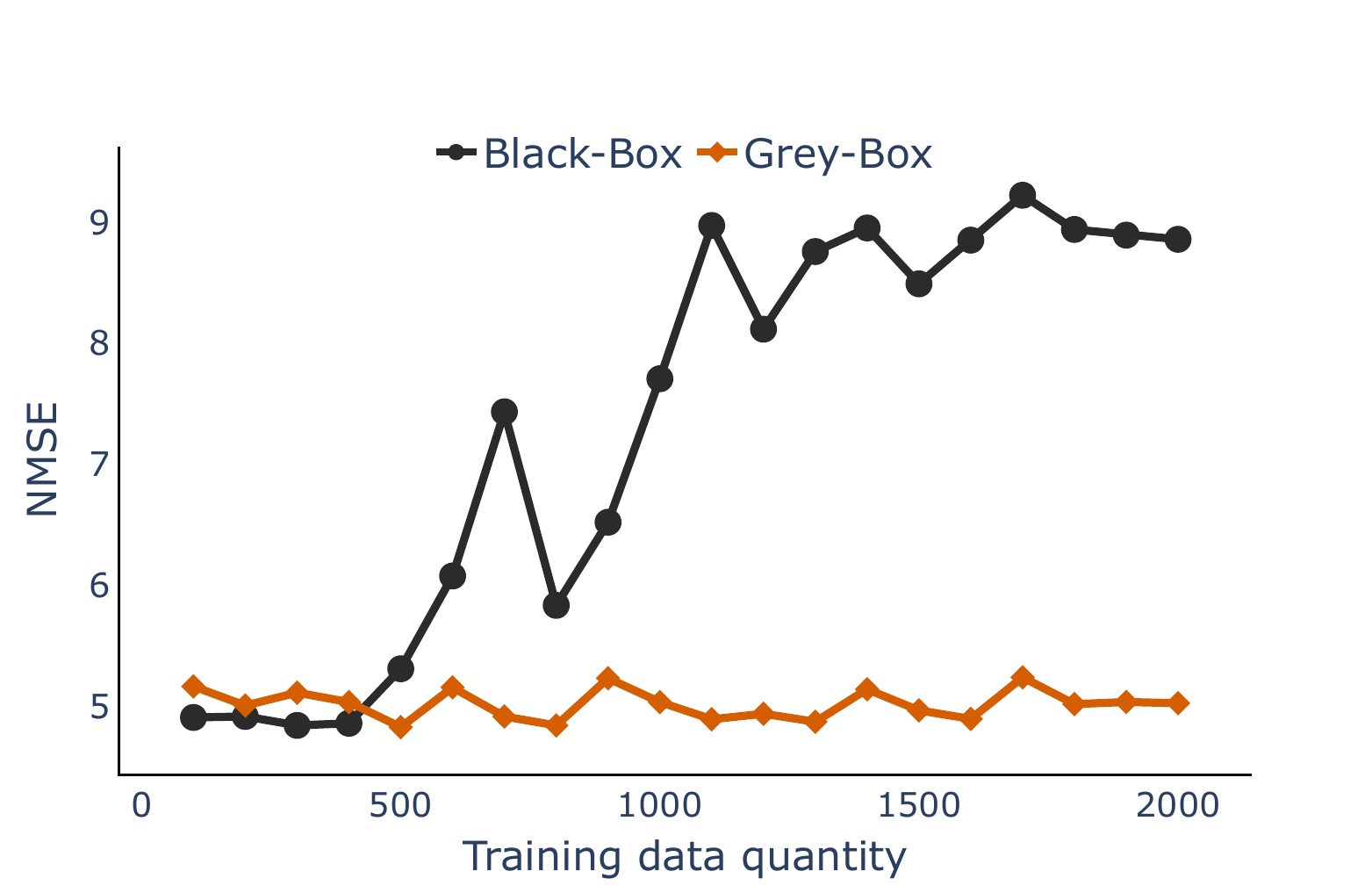}%
        \label{fig:nathan_GP_nmse}%
    }
    \hfill
    \subfloat[Time]{%
        \includegraphics[width=0.48\linewidth]{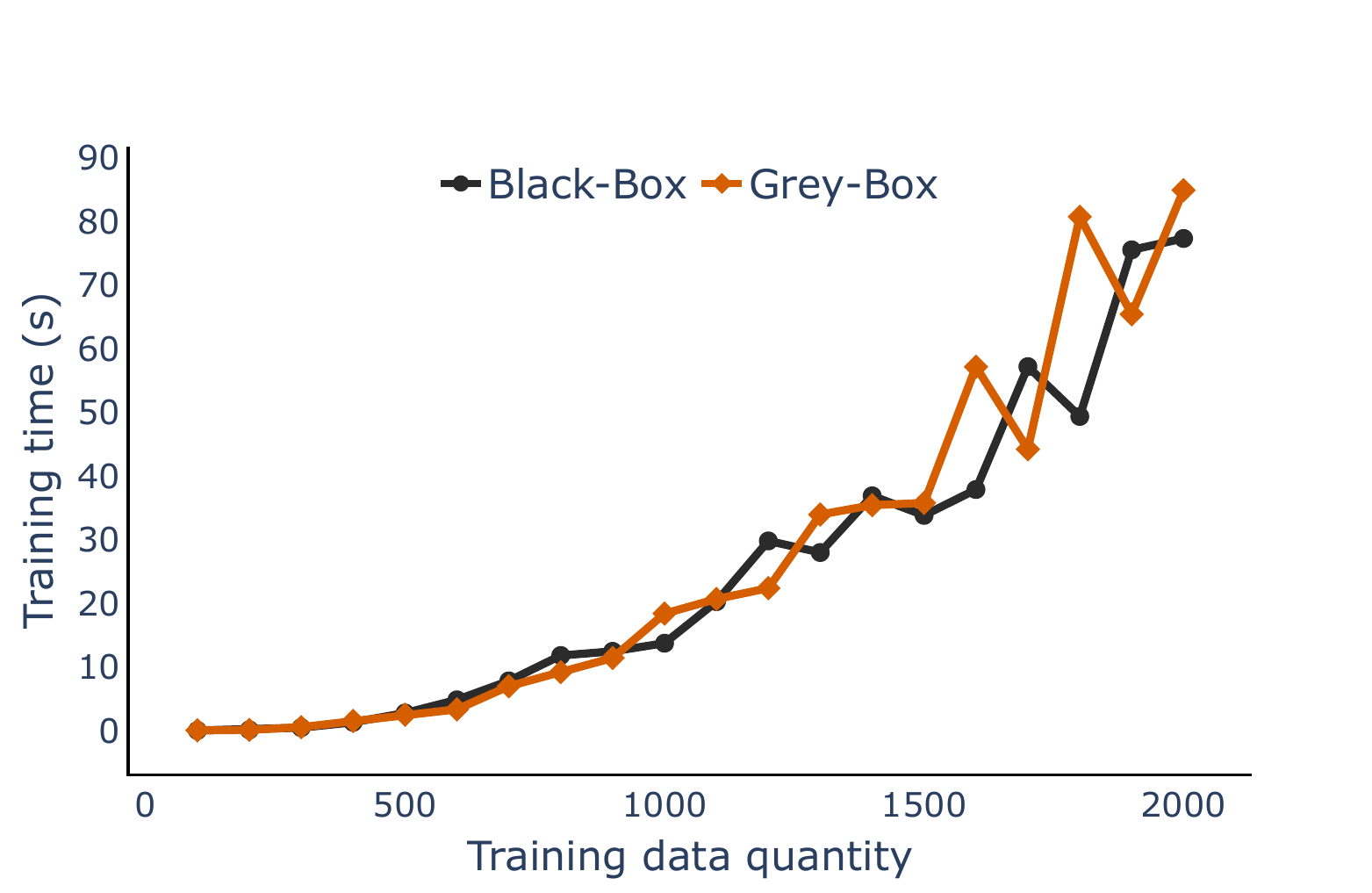}%
        \label{fig:nathan_GP_time}%
    }
    \caption{The NMSE and training time results for the GP input augmentation case study, showing the performance of the black-box and grey-box models.}
    \label{fig:nathan_GP}
\end{figure}

\begin{figure}[h]
    \centering
    \subfloat[NMSE]{%
        \includegraphics[width=0.48\linewidth]{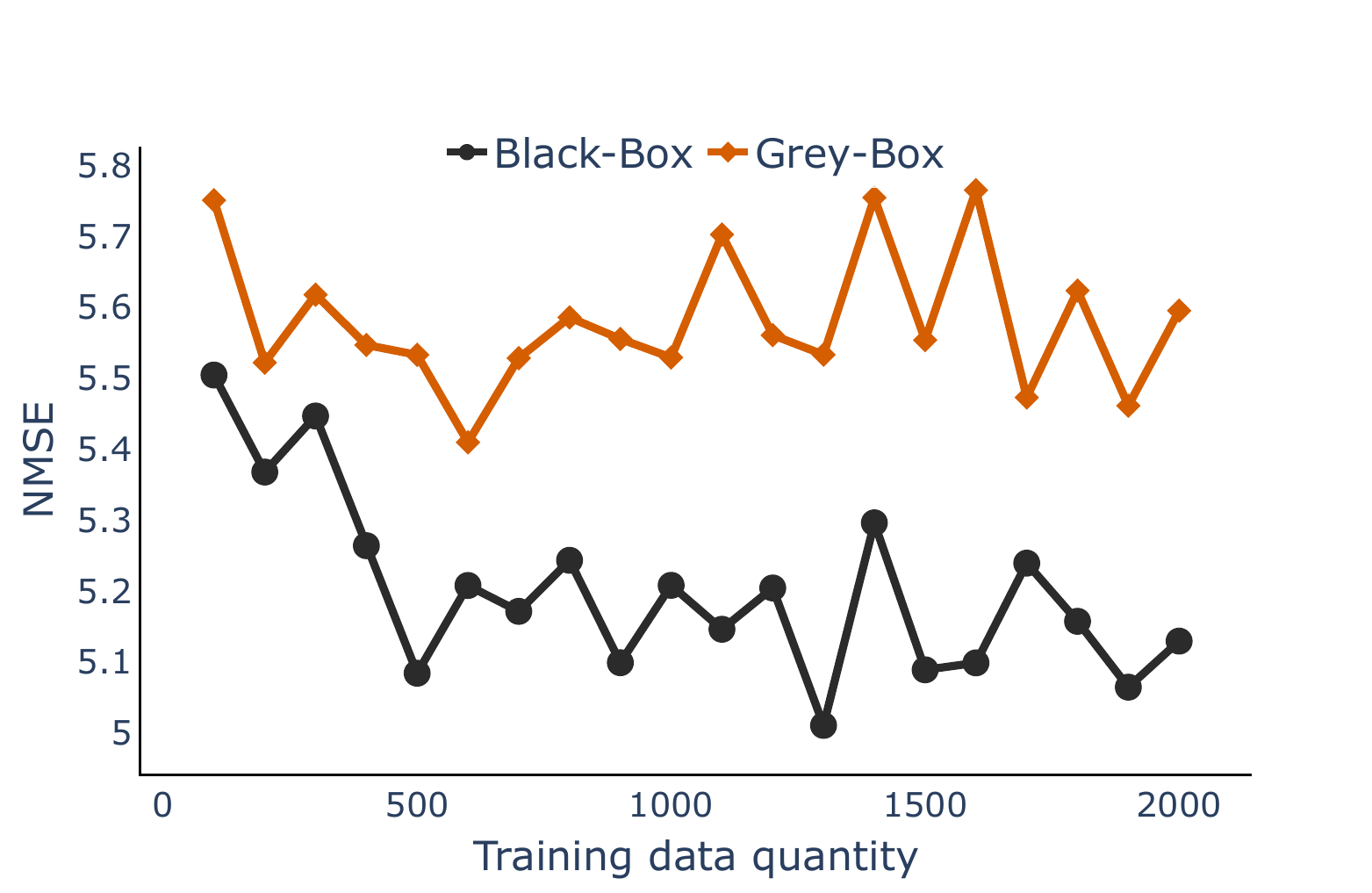}%

        \label{fig:nathan_NN_nmse}%
    }
    \hfill
    \subfloat[Time]{%
        \includegraphics[width=0.48\linewidth]{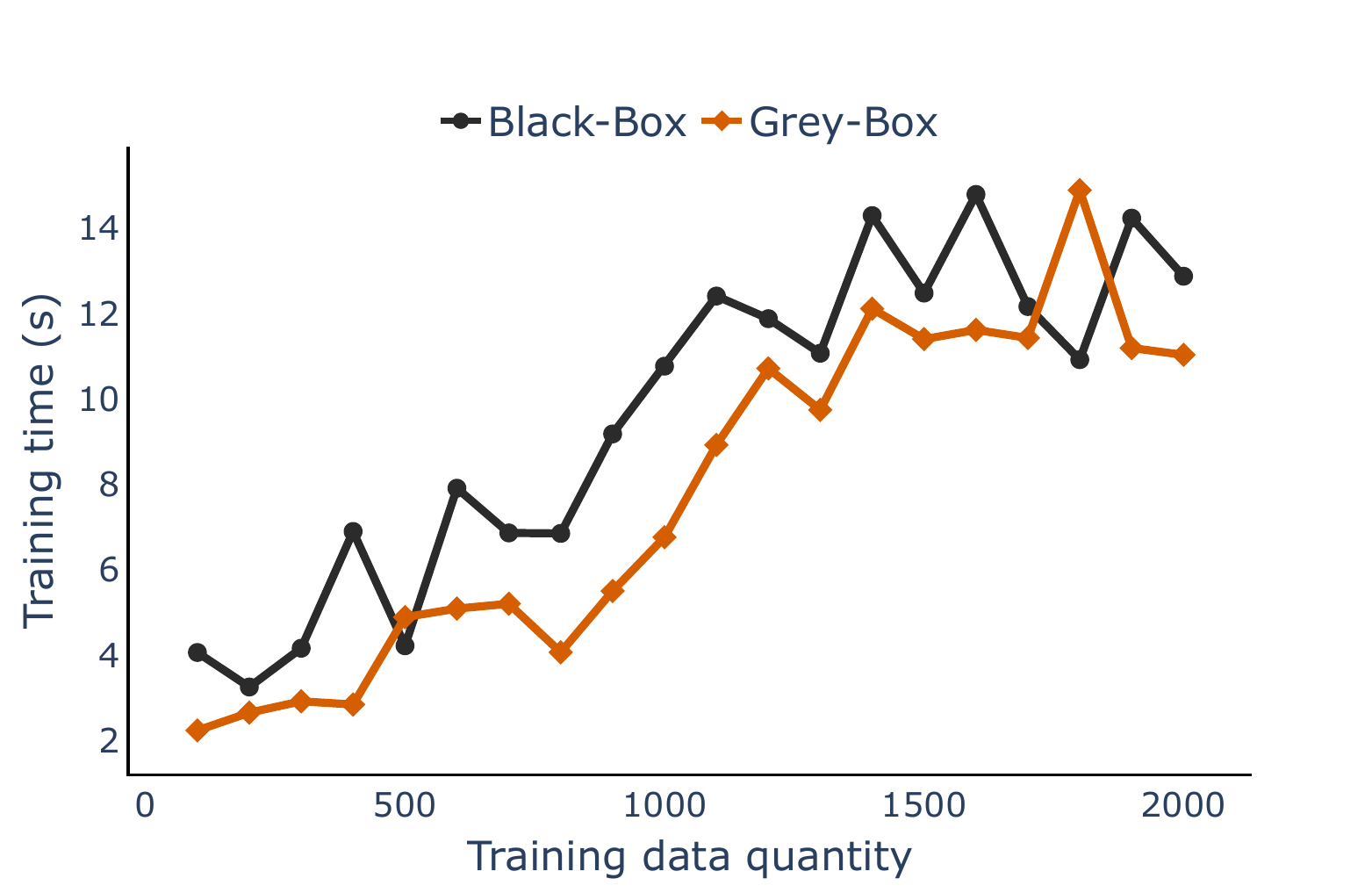}%
        \label{fig:nathan_NN_time}%
    }
    \caption{The NMSE and training time results for the NN input augmentation case study, showing the performance of the black-box and grey-box models.}
    \label{fig:nathan_NN}
\end{figure}

Figure~\ref{fig:nathan_GP} shows the black-box and grey-box GP models' performance and training time results with different training data quantities.
Figure~\ref{fig:nathan_NN} shows the same results for the NN models.

Firstly considering the GP results, shown in Figure~\ref{fig:nathan_GP}, we can see that the black-box has a lower NMSE than the grey-box model at low training data ($<$500 points), before the performance of the black-box model deteriorates while the grey-box remains consistent around an NMSE of 5.
This is likely due to overfitting, where the black-box model learns the 
training data well but fails to generalise to unseen data - a phenomenon that 
is less likely to occur in physics-informed models.
As both models perform well with minimal data, using a performance threshold 
yields no emissions savings for the grey-box model. However, the consistency 
of the grey-box model must be taken into account, as it is more reliable and 
less sensitive to training data quantity.
Figure~\ref{fig:nathan_GP_time} shows that the models take a similar amount of time to train, which is expected as the GP models have the same architecture as each other, using one overall lengthscale rather than one per dimension.

The NN results, shown in Figure~\ref{fig:nathan_NN}, reinforce the notion that input augmentation will not make a carbon saving in this instance.
The models perform similarly, with the black-box model slightly outperforming 
the grey-box model across all training data quantities. 
The training time for the grey-box model is slightly shorter, meaning a small carbon saving could be made here, but the difference is not significant.

The reason why input augmentation does not lead to performance gains in this case study is twofold.
Firstly, some of the black-box inputs show a close to linear relationship 
with strain, such that the black-box problem is already simple to learn 
before augmentation.
On the other hand, while there is a linear relationship between normal force at the CG and strain, the other grey-box inputs do not show a clear correlation with the target, meaning they do not necessarily simplify the problem.
It is important to be aware that there may be times where the data-driven 
model is the best suited to the task, in terms of performance and carbon 
footprint. 
With that said, this is case specific, and in other cases where input augmentation simplifies the relationship between inputs and outputs, input augmentation may lead to carbon savings.
In this case, carbon savings have not been calculated, due to the diverging performance of the black-box GP, and the lack of overlapping performance between the two NNs.

In the next section, we will examine how we can embed knowledge of the physical system into a GP model via our choice of kernel function, and the possible carbon savings this can achieve.
\section{Physics-Informed Kernel Selection}
\label{sec:conference}

\begin{figure}[h]
    \centering
    \includegraphics[width=0.5\linewidth]{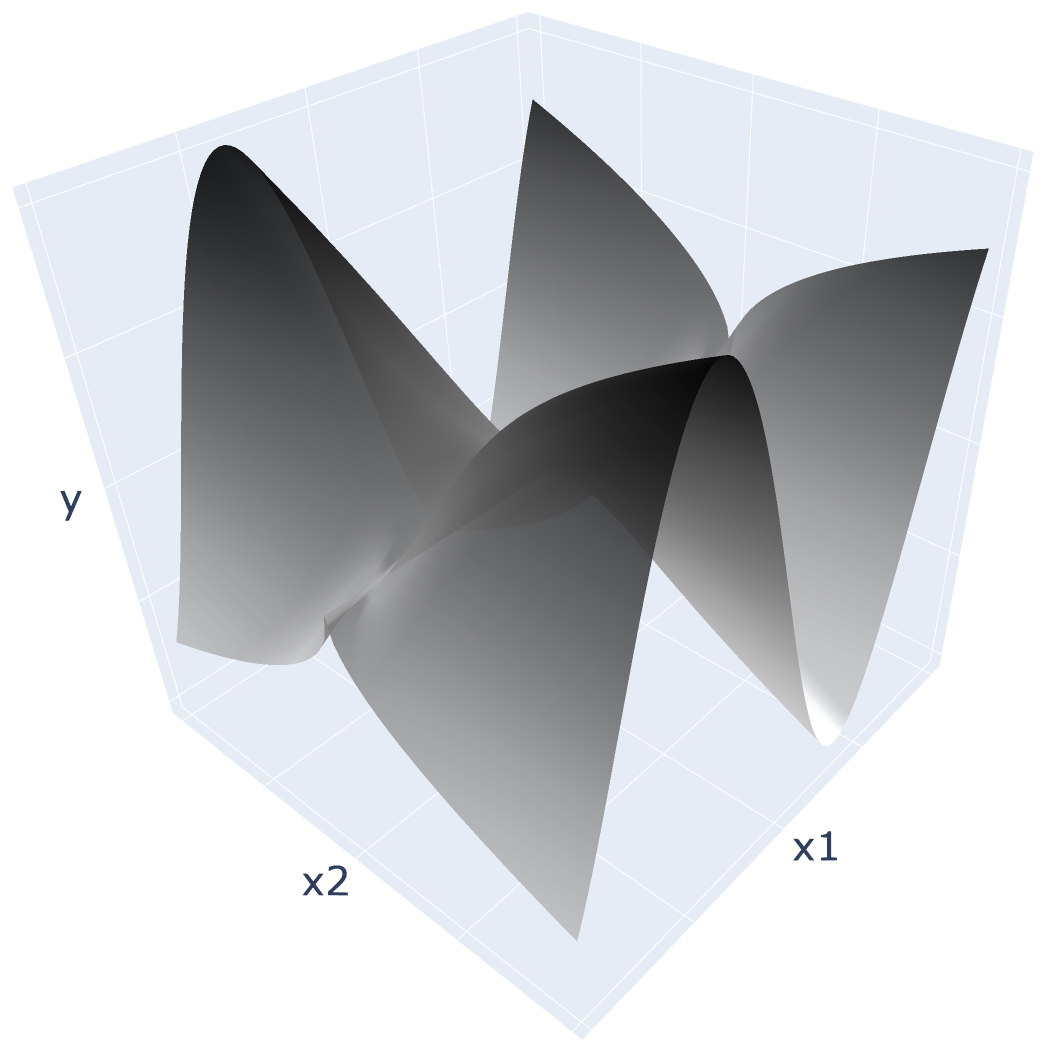}
    \caption{The toy-box surface used to explore physics-informed kernel 
    selection.}
    \label{fig:toybox_surf}
\end{figure}

As outlined in Section~\ref{sec:models}, our choice of kernel function in a GP model can 
be used to incorporate our knowledge of the system's behaviour, using the 
kernel hyperparameters to tune the model to the system. In some cases, the 
optimisation problem can be simplified with this knowledge built in, 
requiring less training data to reach required performance, hereby 
potentially reducing our emissions.
However, some kernel functions, particularly some derived from physics 
\cite{cross2021physics}, are more complex than their more simple black-box 
counterparts, namely that they have more hyperparameters to learn, 
potentially increasing the training time, creating a trade-off between the 
number of hyperparameters and the amount of training data required to reach a 
performance threshold.
To begin to explore this trade-off, a very simple ``toy-box" test case was designed - a periodic surface defined by Equation~\ref{eq:toybox_surface} and shown in Figure~\ref{fig:toybox_surf}. 

\begin{eqnarray}
y = \sqrt{|x_2|}\,\sin\!\left(4\,x_1\right)
    \label{eq:toybox_surface}
\end{eqnarray}

Three GP models were designed with various ``greyness'', with the embedded knowledge concerning the periodicity of the surface. 
Black-1 is a standard black-box GP regression with an SE kernel and a constant mean function, with four hyperparameters to learn - the lengthscale and output variance from the kernel, the mean, and the observation noise.
The two grey-box models differ to Black-1 by multiplying the SE kernel by a 
periodic kernel, also known as the "locally-periodic" kernel \cite{duvenaud2014automatic}, outlined in Equation~\ref{eq:periodic_kernel}, which we use 
as our physics knowledge proxy.

\begin{eqnarray}
k_{Per}(x,x') = \sigma^2exp(-\frac{2sin^2(\pi |x-x'|/p)}{l^2})
    \label{eq:periodic_kernel}
\end{eqnarray}
 
The periodic kernel consists of 3 hyperparameters - lengthscale ($l$), output 
variance ($\sigma^2$), and period ($p$).
The grey-box models have a total of 6 hyperparameters - the lengthscale and period from the periodic kernel, the lengthscale from the SE kernel, an overall output variance, the mean, and the observation noise.
To mimic scenarios of more or less information available on system parameters, in the first grey-box model, Grey-1, the period is bounded within 10\% of the true value, and in the second grey-box model, Grey-2, the period is fixed at the true value and therefore not optimised in the training process.
Therefore, Grey-2 is ``whiter'' than Grey-1, and has one fewer hyperparameter 
to learn. 
A description of the three models is shown in Table~\ref{tab:model_details}.

\begin{table}[h]
  \centering
  \small
  \setlength{\tabcolsep}{4pt}
  \renewcommand{\arraystretch}{1.1}
  \caption{Model details for Black-1, Grey-1, and Grey-2.}
  \begin{tabular}{lccc}
    \toprule
    & Black-1 & Grey-1 & Grey-2 \\
    \midrule
    Mean function & Constant & Constant & Constant \\
    Covariance function & SE & SE*Periodic & SE*Periodic \\
    Number of hyperparameters to be optimised & 4 & 6 & 5 \\
    \bottomrule
  \end{tabular}

    \label{tab:model_details}
\end{table}

The models predicted the amplitude of the surface, y, given $x_1$ and $x_2$ 
inputs. 
Coverage was varied in 10\% intervals across the $x_1$ variable, with 100 training data points per 10\%.
To train the models, the ADAM gradient descent optimiser \cite{kingma2017adammethodstochasticoptimization} was used with 1000 iterations.
While the footprints of these models will be very small, this work aims to form the foundation of a concept that can be scaled up to much larger models, where carbon savings become much more significant.

\subsection{An engineering test case}

\begin{figure}[htbp]
    \centering
    \includegraphics[width=0.75\linewidth]{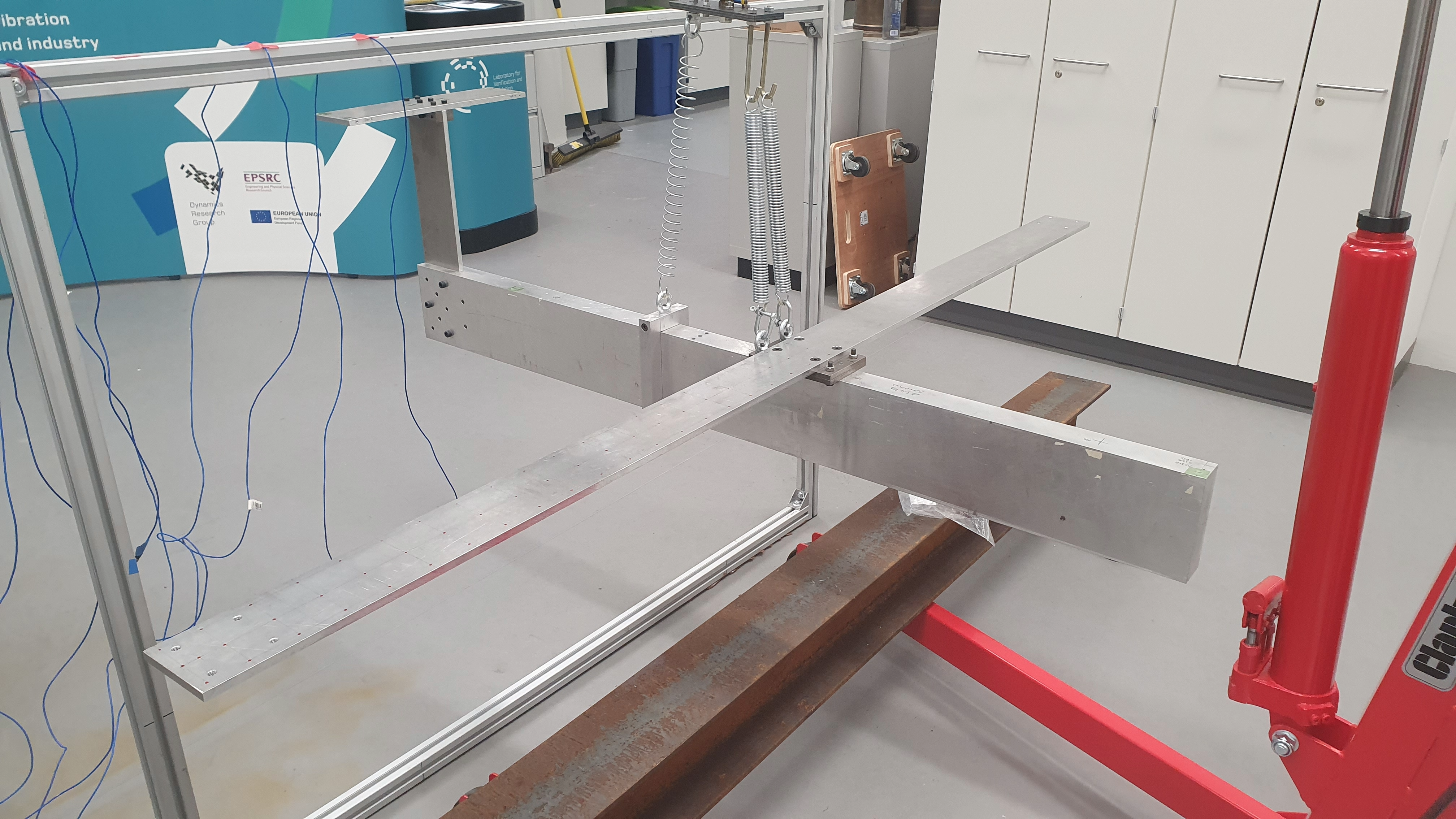}
    \caption{The GARTEUR structure used to collect the engineering test case 
    data.}
    \label{fig:GT_photo}
\end{figure}

The same methodology described above was applied to a set of engineering 
data, collected for use in an SHM study from a lab-based modal test of a 
simple metallic aircraft structure - the GARTEUR SM-AG19 benchmark, shown in 
Figure~\ref{fig:GT_photo} (see \cite{Link2003} and \cite{Gibson2024PIML} for 
specific details). 

\begin{figure}[H]
\centering

\subfloat[Original]{
    \includegraphics[width=0.4\textwidth]{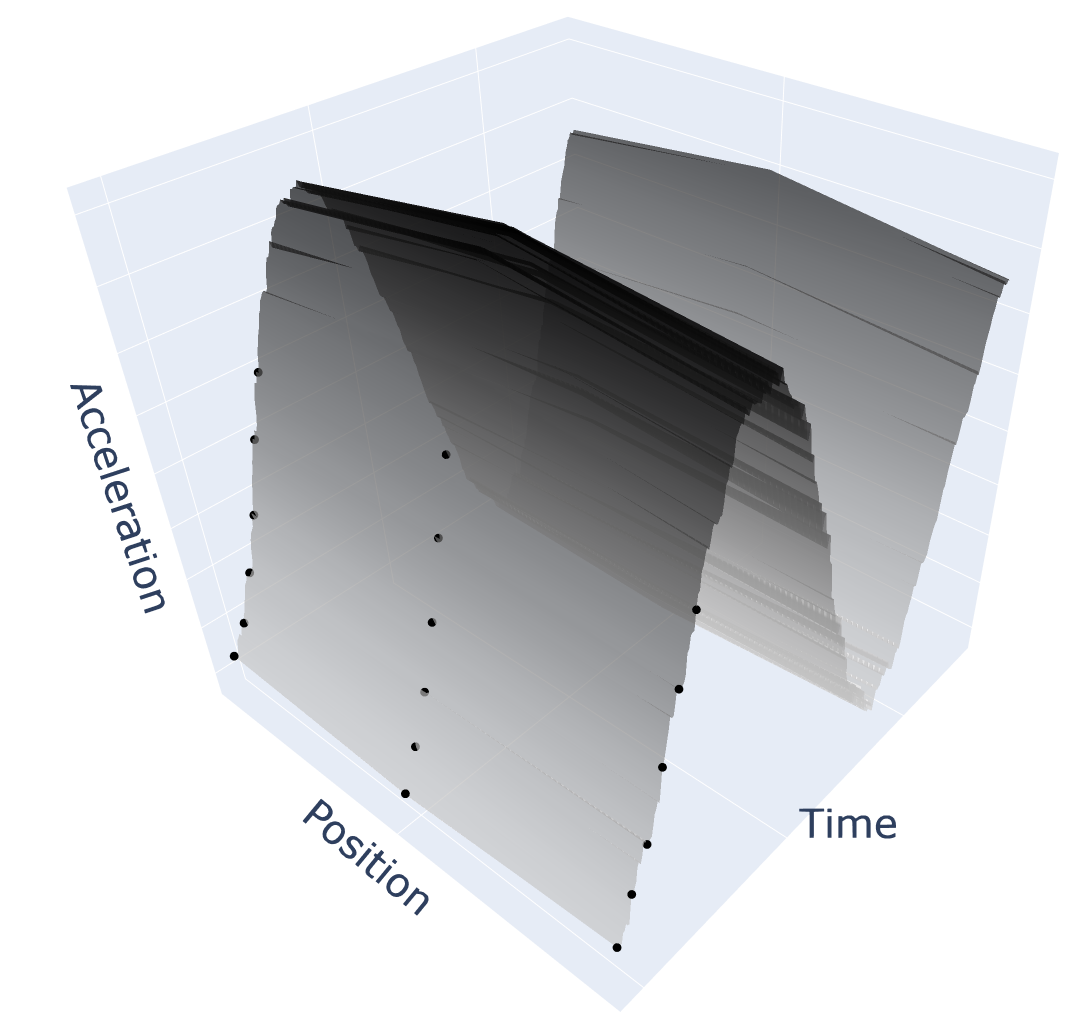}}
\hspace{0.02\textwidth}
\subfloat[Upsampled]{
    \includegraphics[width=0.4\textwidth]{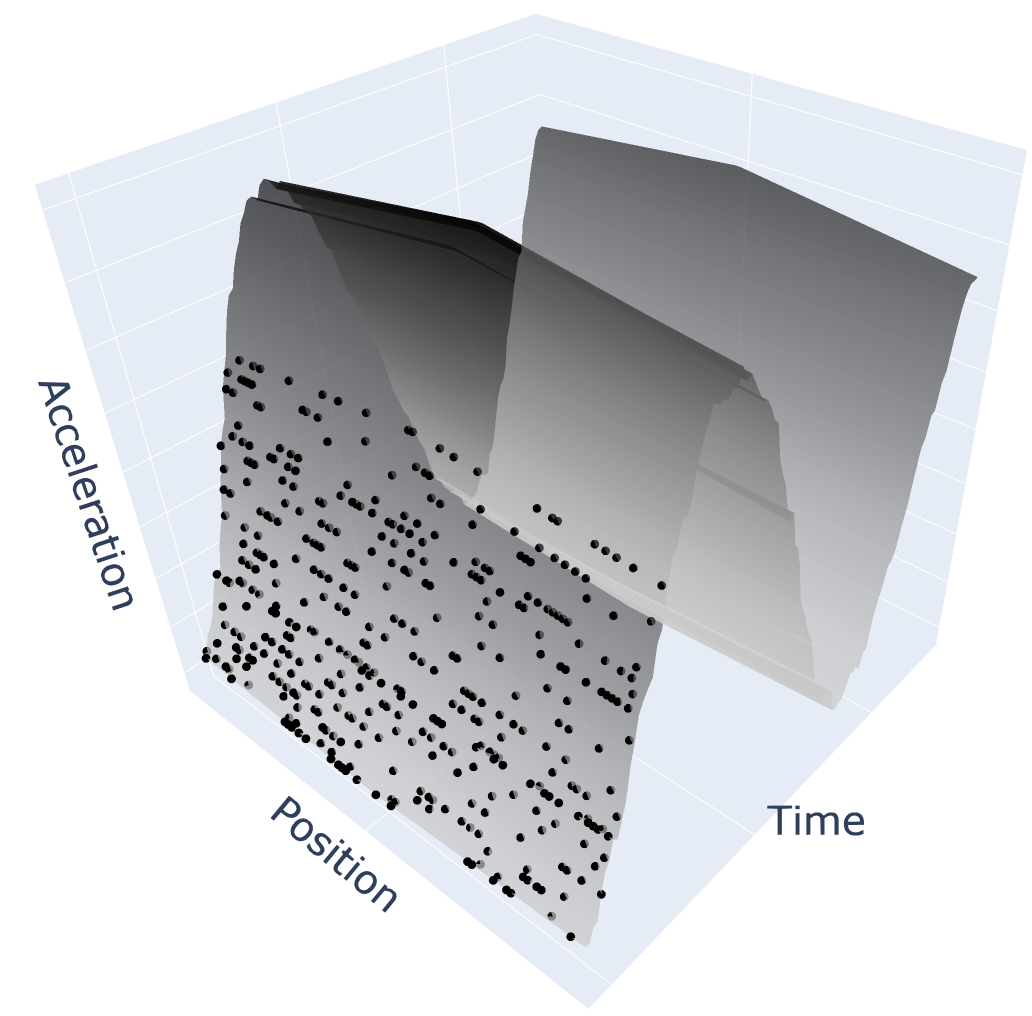}}

\caption{The different training data layouts used for the GARTEUR test cases 
with 20\% training data coverage.}
\label{fig:GT_surf}
\end{figure}

For this work, the wing, instrumented with accelerometers, was excited via an 
electrodynamic shaker to produce a response in its first vibrational mode (close to 
10Hz), producing a similar learning surface to that of the toy-box example 
above - see Figure~\ref{fig:GT_surf}. 
The ML task for this initial study is to predict the acceleration response spatially and temporally, with the same periodic kernel employed as above used to bring in ``knowledge" of the periodic nature of the modal response. 
Two test cases were employed: the first using the accelerometer positions and every tenth data point, the second using a spatially upsampled dataset, as can be seen in Figure~\ref{fig:GT_surf}, such that each 10\% of surface coverage contains 200 training data points.

\subsection{Results and Discussion}
\begin{figure}[!t]
    \centering
    \subfloat[NMSE]{%
        \includegraphics[width=0.48\linewidth]{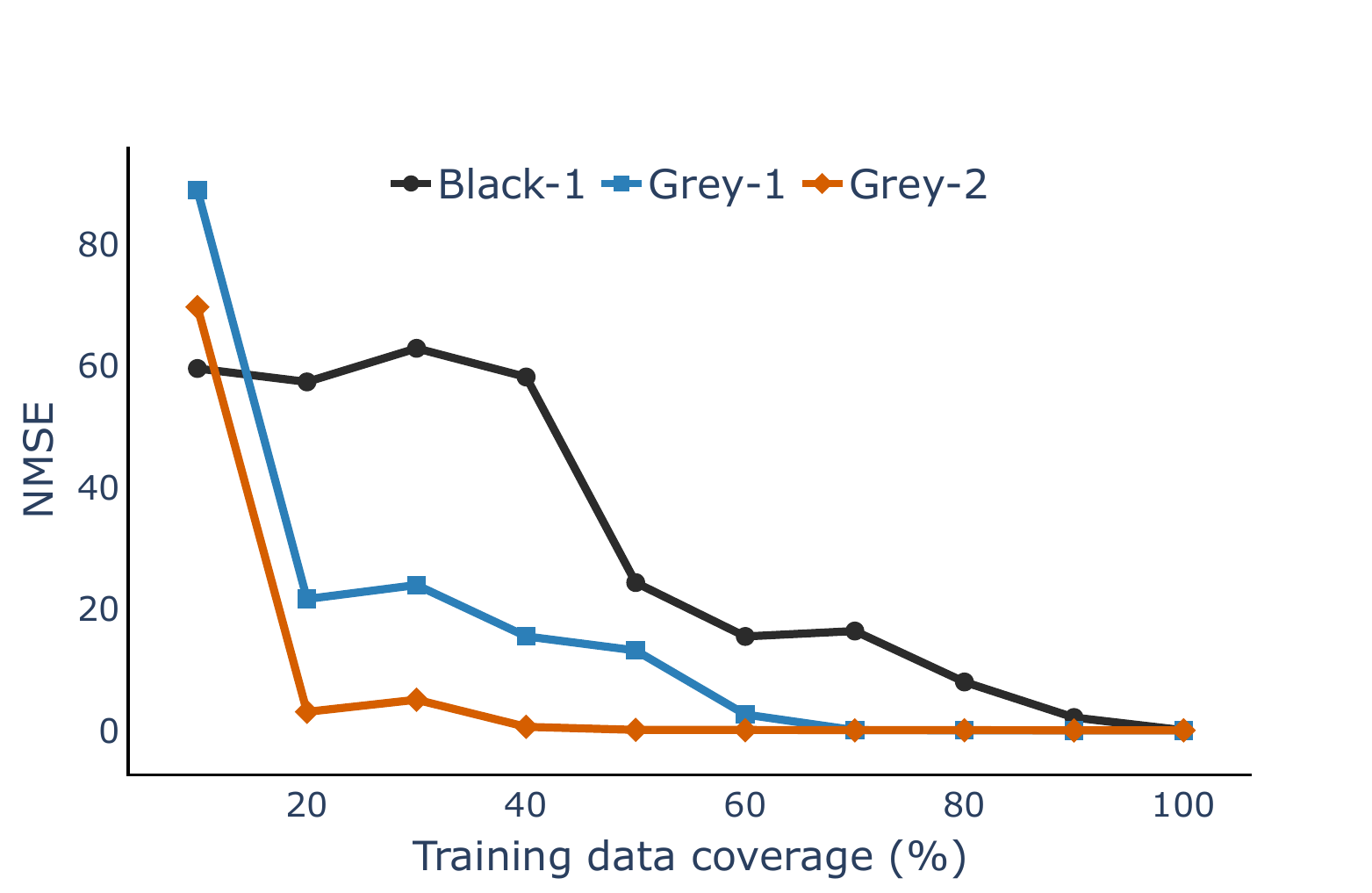}%
        \label{fig:toybox_nmse}%
    }
    \hfill
    \subfloat[Time]{%
        \includegraphics[width=0.48\linewidth]{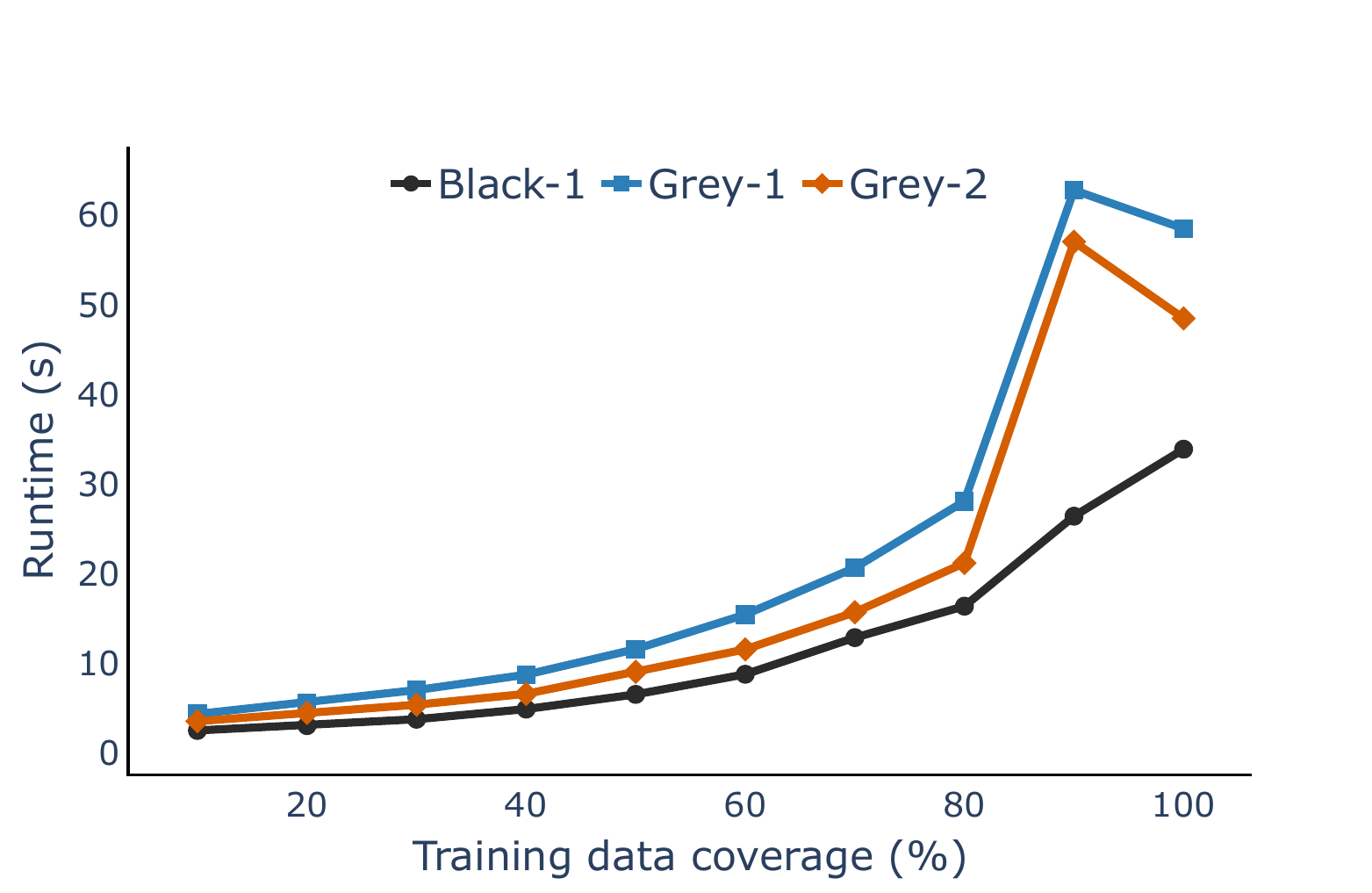}%
        \label{fig:toybox_time}%
    }
    \caption{Toy-box NMSE and run time results.}
    \label{fig:toybox_results}
\end{figure}

\begin{figure}[!t]
    \centering
    \subfloat[NMSE]{%
        \includegraphics[width=0.48\linewidth]{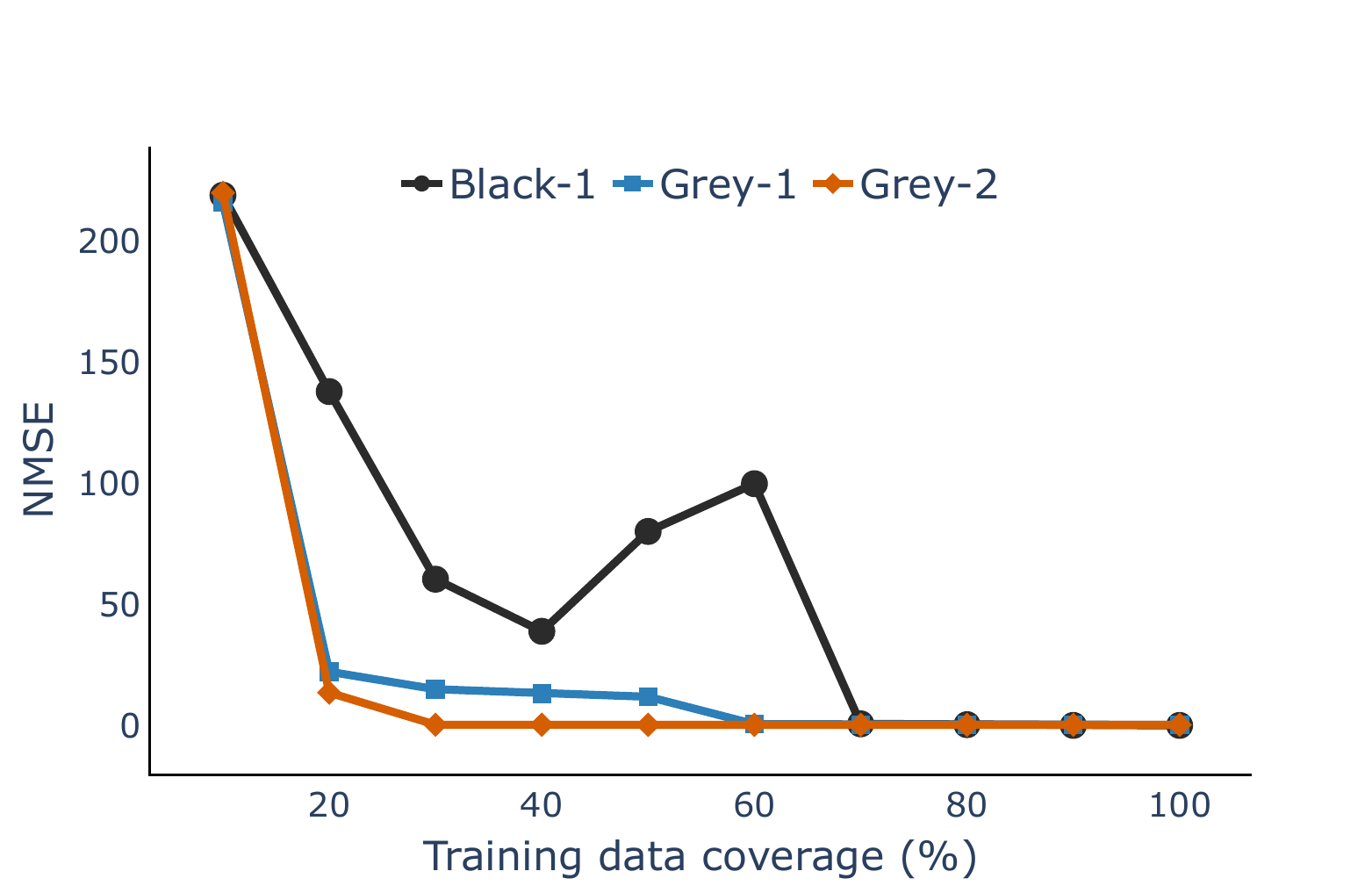}%
        \label{fig:GT_original_nmse}%
    }
    \hfill
    \subfloat[Time]{%
        \includegraphics[width=0.48\linewidth]{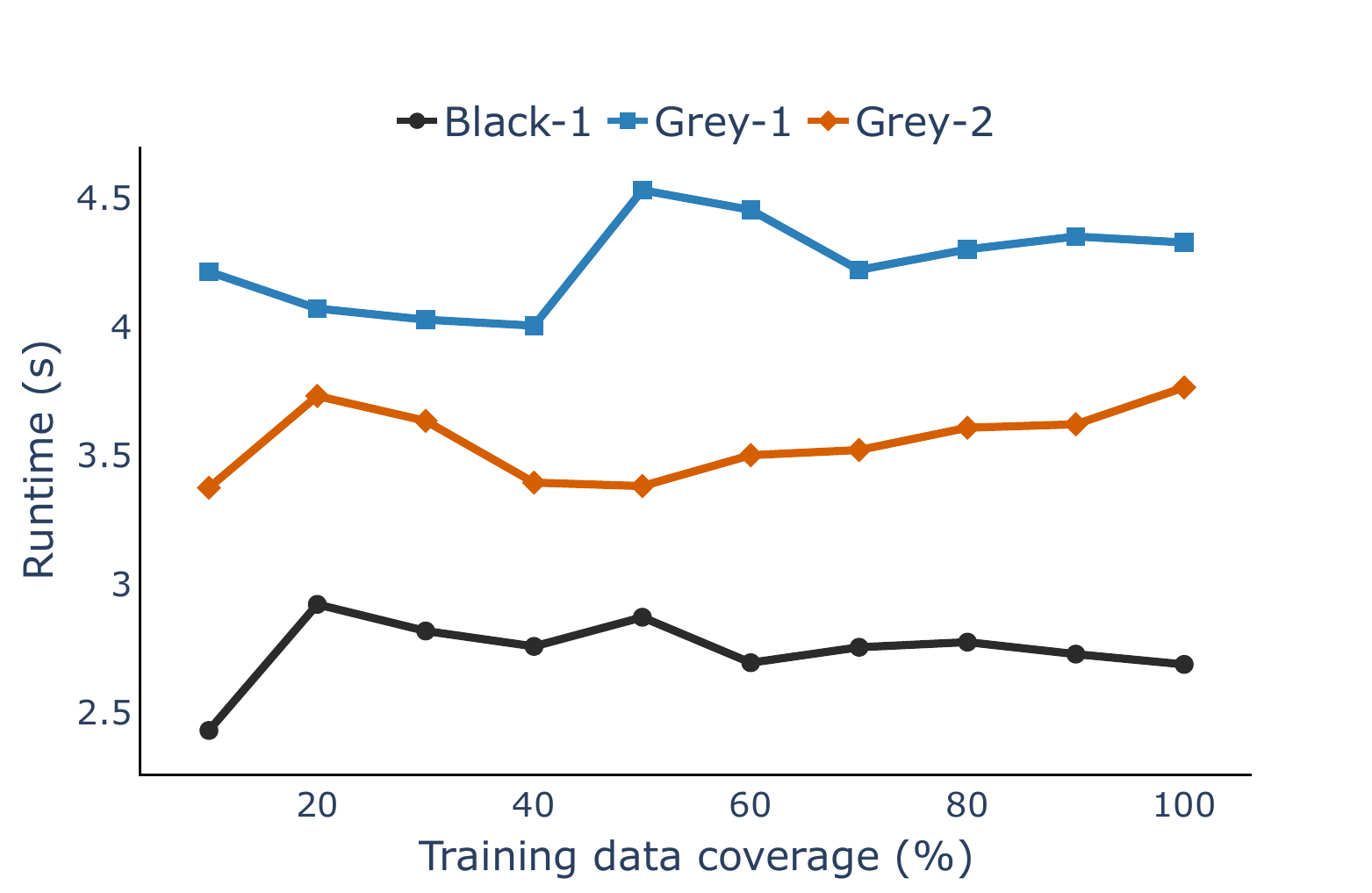}%
        \label{fig:GT_original_time}%
    }
    \caption{NMSE and run time results for the original GARTEUR data layout.}
    \label{fig:GT_original_results}
\end{figure}

\begin{figure}[!t]
    \centering
    \subfloat[NMSE]{%
        \includegraphics[width=0.48\linewidth]{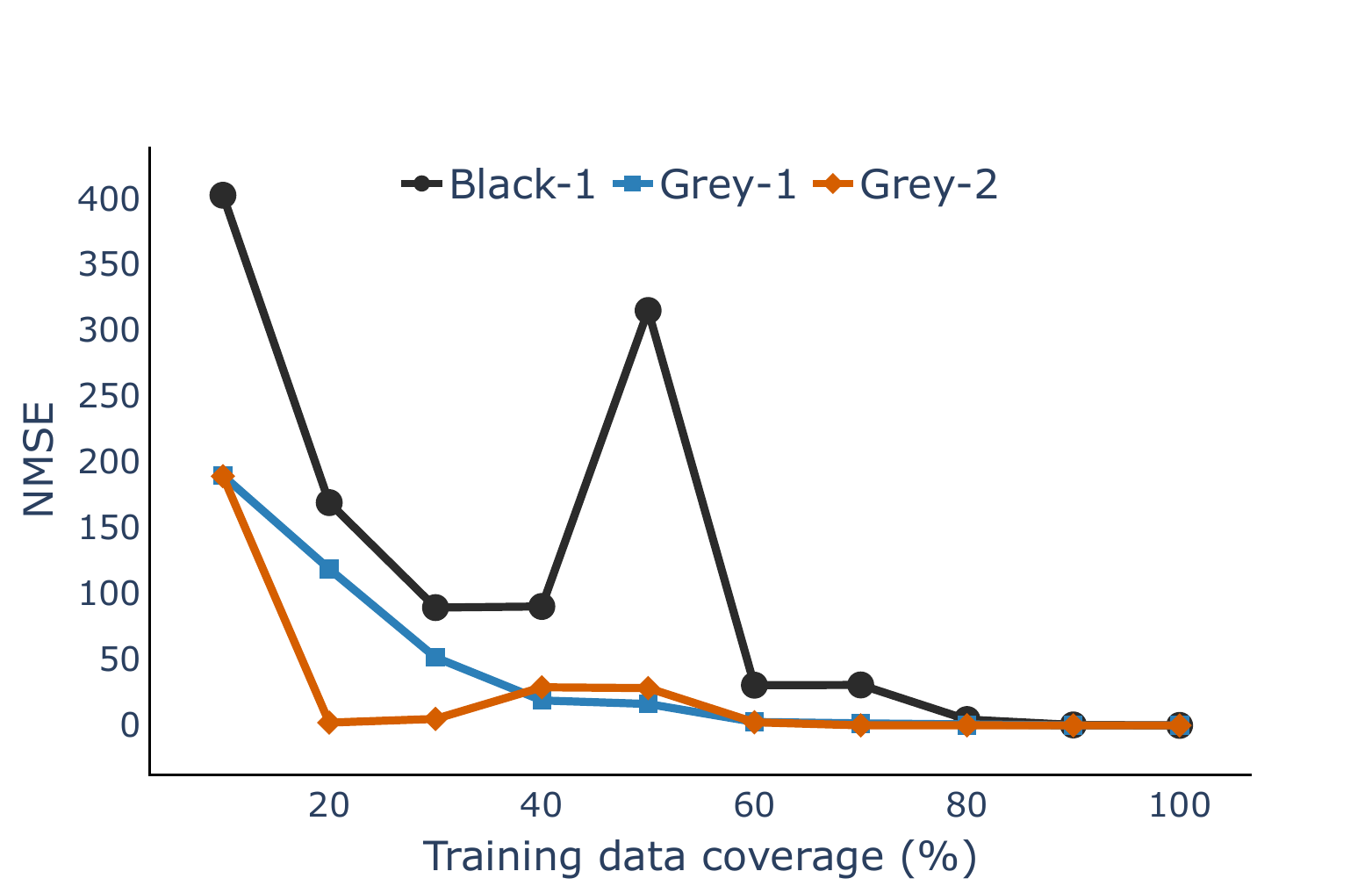}%
        \label{fig:GT_upsampled_nmse}%
    }
    \hfill
    \subfloat[Time]{%
        \includegraphics[width=0.48\linewidth]{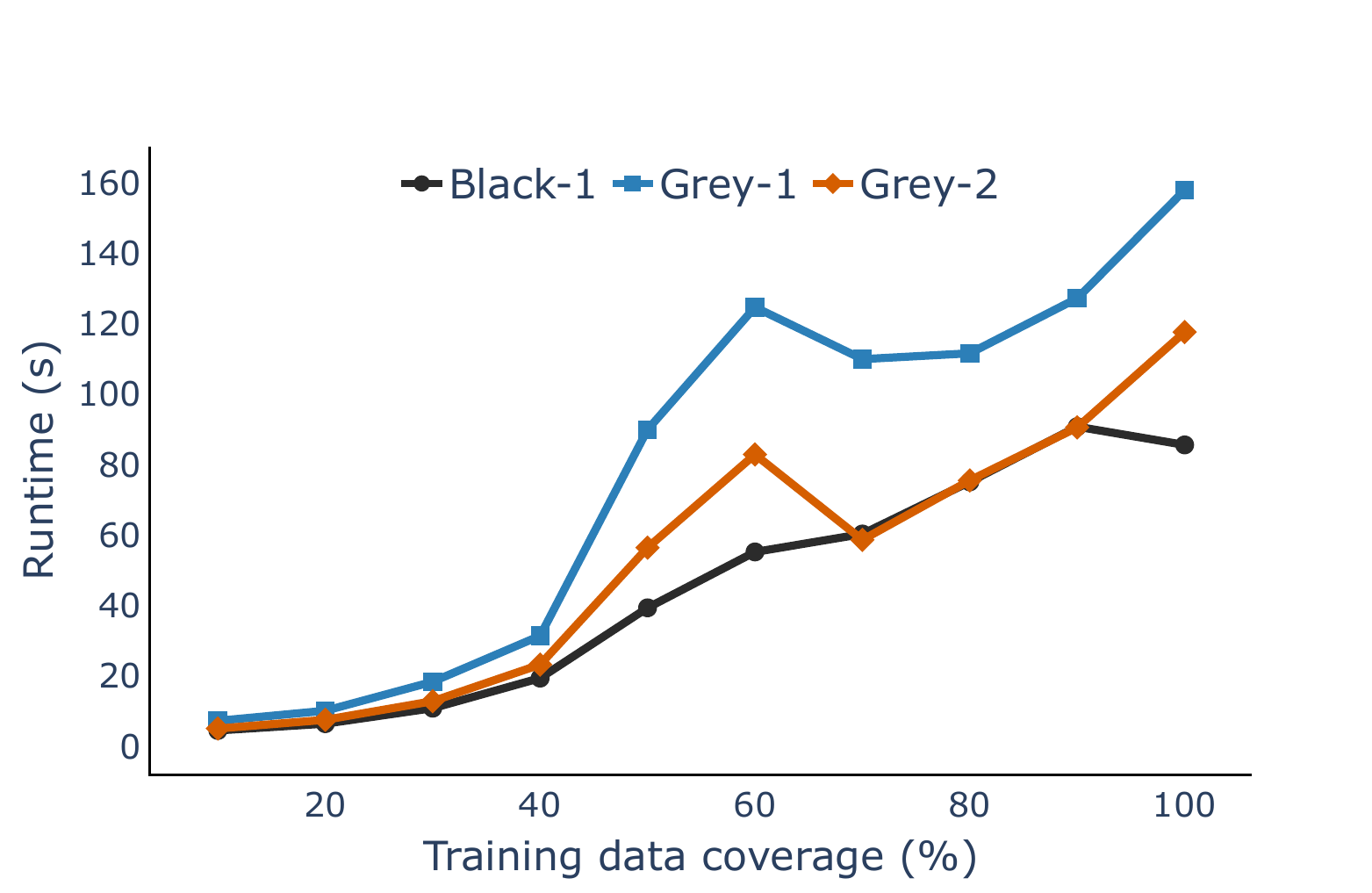}%
        \label{fig:GT_upsampled_time}%
    }
    \caption{NMSE and run time results for the upsampled GARTEUR data layout.}
    \label{fig:GT_upsampled_results}
\end{figure}

Figures~\ref{fig:toybox_results}, \ref{fig:GT_original_results} and 
\ref{fig:GT_upsampled_results} show the NMSE and run time results of the 
grey-box and black-box models for each of the three test cases. These show 
that the NMSE of the physics-informed models drops sharply compared to the 
black-box with increasing coverage. As a result, the grey-box models would 
require less training data to meet an NMSE threshold compared to the 
black-box, with Grey-2 requiring the least, as expected due to the additional 
knowledge embedded in the model.

In terms of training time, Figures ~\ref{fig:toybox_time}, 
\ref{fig:GT_original_time} and \ref{fig:GT_upsampled_time} show that for a 
given coverage, the black-box has the fastest run time, followed by Grey-2, 
with Grey-1 being the slowest, correlated to the number of hyperparameters in 
each model.
In a Gaussian process regression we expect that runtime is dominated by a term proportional to the training data size ($N$) cubed, with a secondary term from the gradient descent proportional to the number of hyperparameters to be optimised ($H$) and training data squared \cite{RasmussenWilliams2006}, making the computation $\sim \mathcal{O}(N^3)+\mathcal{O}(HN^2)$.
In a standard GP implementation, we therefore expect that when $N$ is large, the impact of the hyperparameter optimisation on runtime is insignificant, though it is more prominent at small $N$, especially in complex models with many hyperparameters.
In this initial exploratory work both hyperparameter counts and training data sizes are small in themselves.

Taking an NMSE threshold of $<$10, we can see that in each case, Grey-2 can achieve this with the least training data, followed by Grey-1, and then Black-1.
In the toy-box case, this leads to emissions reductions of 73.3\% and 6.7\% when comparing Grey-2 and Grey-1 to Black-1 respectively (0.45 gCO$_2$e for Black-1, 0.42 gCO$_2$e for Grey-1, and 0.12 gCO$_2$e for Grey-2), showing that the increase in model complexity from additional hyperparameters is overcome by the runtime reduction from reducing training data.
Conversely, considering the original GARTEUR data, Grey-1 and Grey-2 increase the emissions by 71.4\% and 42.9\% respectively, compared to Black-1 (0.07 gCO$_2$e for Black-1, 0.12 gCO$_2$e for Grey-1, and 0.10 gCO$_2$e for Grey-2).
Finally, the upsampled GARTEUR case shows Grey-1 increase the emissions by 65.7\% and Grey-2 reduce them by 89.7\% compared to Black-1 (2.04 gCO$_2$e for Black-1, 3.38 gCO$_2$e for Grey-1, and 0.21 gCO$_2$e for Grey-2).
These sets of results highlight that using physics to reduce training data 
requirements is not enough in itself to reduce the emissions of a model - we 
must also consider the complexity of the optimisation, here driven by 
hyperparameter count.
As expected, the results also suggest that the emissions savings from 
introducing physics will be more prominent at large training data sizes 
(which offers a reduction in emissions from collecting and storing data as 
well as runtime).

The next section will investigate a case study where the kernel itself has 
been designed to include physics knowledge in the form of a switching model, 
adding increased complexity again into the grey-box learner.
\section{Switching Kernels}
\label{sec:Dan}
As a second example of a hybrid model, where the explanatory power of the 
data-driven and physics components are more equally shared than other 
approaches, we explore here the performance of GPs with and without switching 
behaviour built into the covariance function. Switching kernels are helpful 
when we see a change in the system behaviour under certain conditions.
This change can be difficult for a black-box model to learn, particularly if 
there is little data from some operating regimes, giving the grey-box model 
in this case study a significant performance advantage.
This is somewhat unusual - in most cases, physics-informed methods aim to restrict model flexibility or capabilities to encourage physically plausible predictions.
Here however, the switching kernel structure expands the range of behaviours that the model can predict.
Given the increased 
performance, we might expect to see significant gains in reducing run time, 
however, the switching kernels themselves have more hyperparameters to learn 
compared to a standard SE kernel, and the optimisation problem can become 
more complex.
Once again, we must explore the trade-off between optimisation complexity and 
training data requirements to determine if this PIML method can reduce 
emissions.

The models used in this case study were first developed in \cite{pitchforth2026physics} to investigate the deck acceleration of the Tamar Suspension Bridge in response to wind load and direction.
As the bridge is orientated in the east-west direction, high speed winds from the north and south will cause higher deck accelerations than winds from the east and west \cite{cross2012structural}.
The data shows a quadratic relationship between high speed southerly winds and deck acceleration, suggesting that these winds generate lift on the bridge deck.
To encapsulate this behaviour, a second order polynomial kernel was used to 
represent the lift force.
However, this force is only present under certain loading conditions, so a 
composite weighting function with two sigmoids was used to ensure the lift 
kernel only activates when the wind speed is high and aligns with the 
north-south direction.
The final physics-informed kernel mixture is shown in Equation~\ref{eq:dan_kernel} \cite{pitchforth2026physics}.

\begin{equation}
\begin{aligned}
k(\theta,\theta',u,u') ={}&
\underbrace{
w_{\mathrm{comp}}(u,u',\theta,\theta')
w_{\mathrm{comp}}(u,u',\theta,\theta')
k_{\mathrm{Lift}}(u,u')
}_{\substack{\text{lift force at high N/S winds}}}
\\[4pt]
&+
\underbrace{
w_{\mathrm{comp}}(-u,-u',-\theta,-\theta')
w_{\mathrm{comp}}(u,u',\theta,\theta')
k_{\mathrm{SE}}(u,u')
}_{\substack{\text{otherwise, use a Squared}\\
\text{Exponential kernel}}}
\end{aligned}
\label{eq:dan_kernel}
\end{equation}

\noindent where $u$ is wind speed, $\theta$ is wind angle, $w_{comp}$ is the 
composite 
weighting function, $k_{Lift}$ is the second order polynomial kernel and 
$k_{SE}$ is the SE kernel.
Additionally, a regime-dependent heteroscedastic noise term was added to incorporate the increased variance in deck acceleration at high wind speeds.
In \cite{pitchforth2026physics}, several models with varying ``greyness" were compared in terms of their performance with 300 randomly selected training points up to 30 mph winds to demonstrate their extrapolation ability.
Here, we use the SE kernel model as our black-box model, and the 
physics-informed mixture kernel with linearly scaling heteroscedastic noise 
as our grey-box model, varying the training data between 100 and 1000 points 
at intervals of 100, maintaining the same 30 mph limit in training.
The optimiser is a QPSO with a population size of 200 and a tolerance of $10^{-4}$ and three restarts.

\subsection{Results and Discussion}

\begin{figure}[h]
    \centering
    \subfloat[NMSE]{%
        \includegraphics[width=0.48\linewidth]{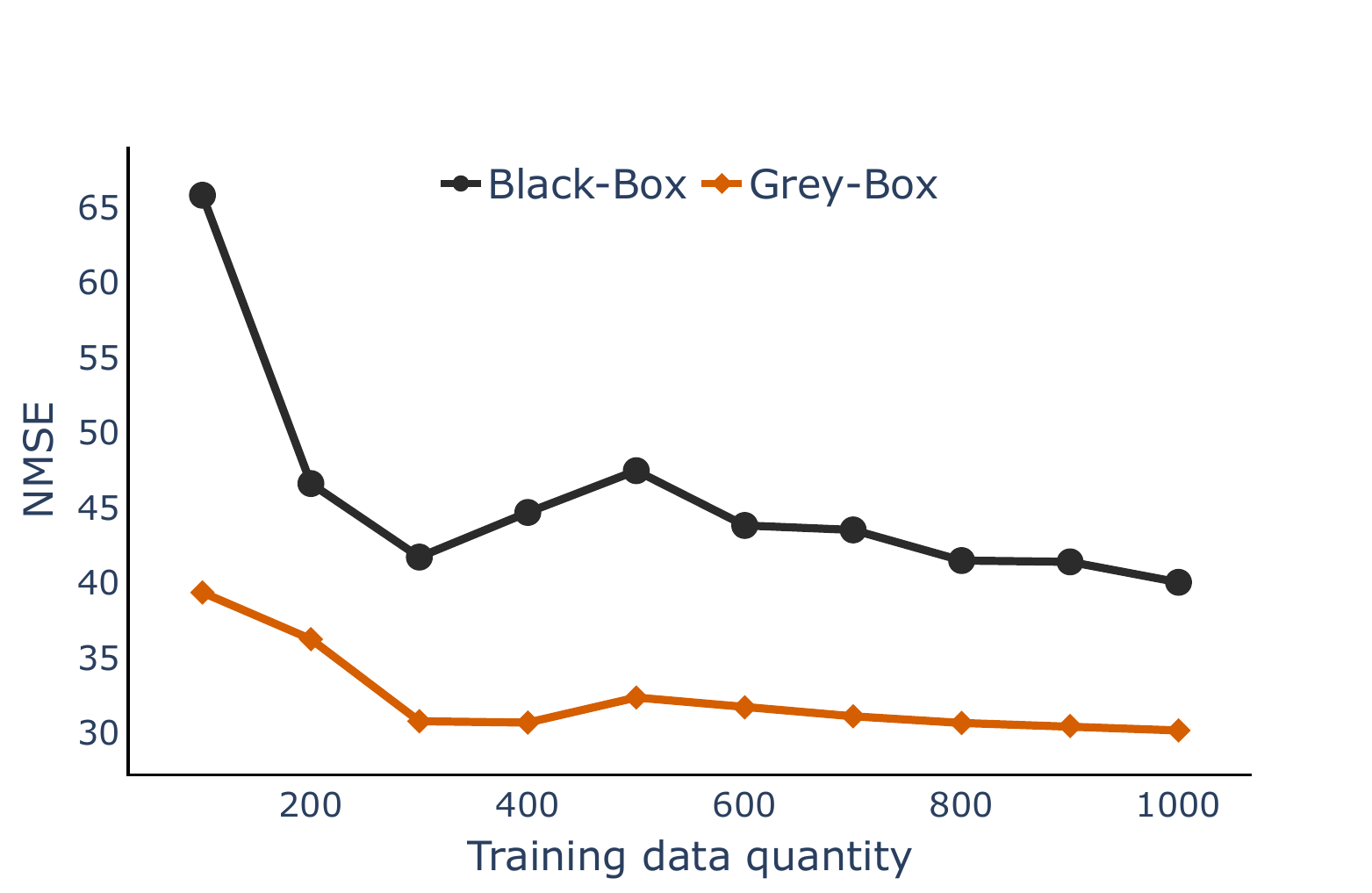}%
        \label{fig:dan_nmse}%
    }
    \hfill
    \subfloat[Training Time]{%
        \includegraphics[width=0.48\linewidth]{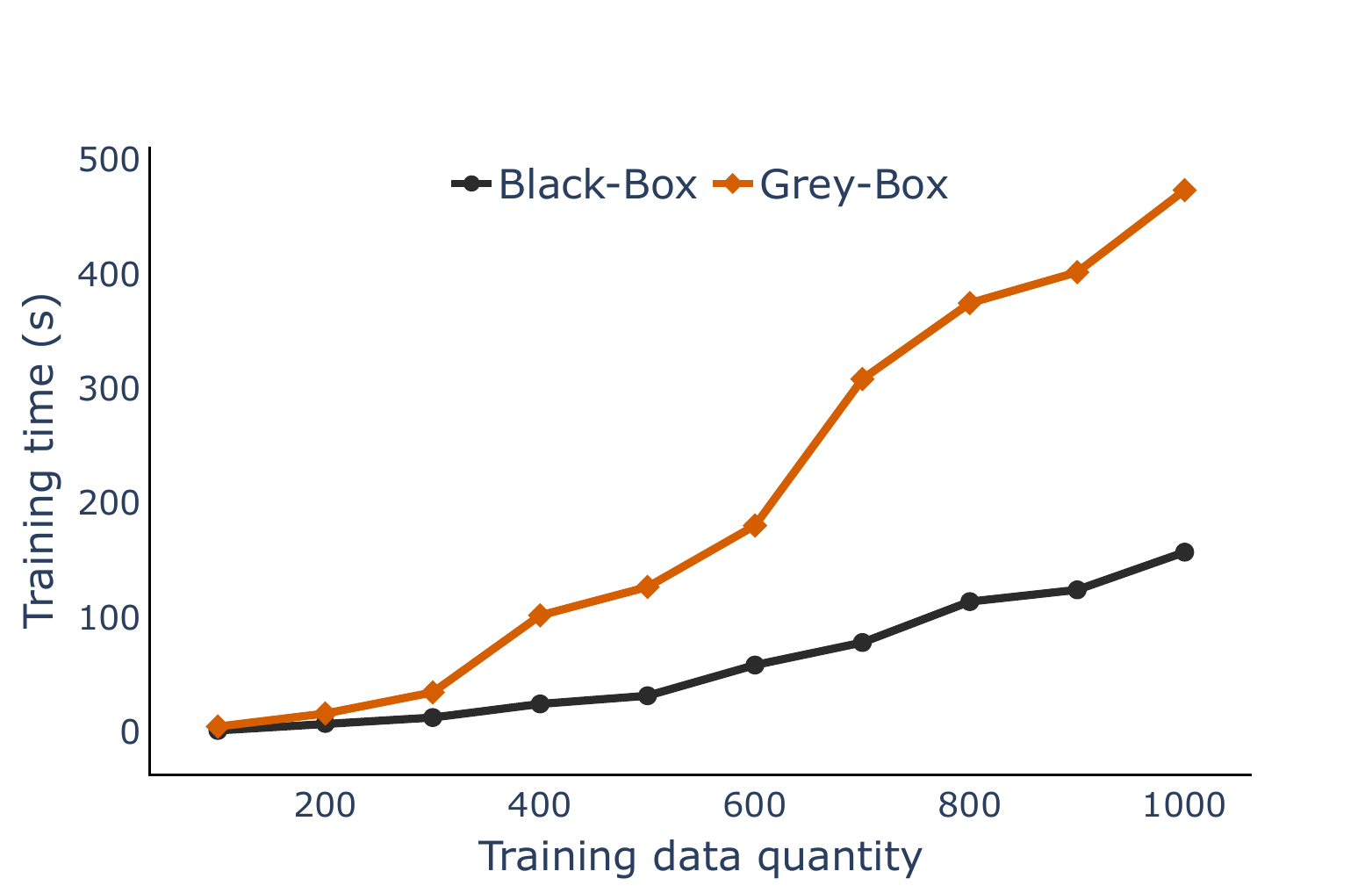}%
        \label{fig:dan_time}%
    }
    \caption{The results from the switching kernel case study, showing the NMSE and training time for the black-box and grey-box models.}
    \label{fig:dan_results}
\end{figure}

Figure~\ref{fig:dan_results} shows the NMSE and training time results for the black-box and grey-box models with different training data quantities.

Figure~\ref{fig:dan_nmse} shows that the grey-box model consistently outperforms the black-box model across all training data quantities and the model performances do not converge.
This is to be expected - for a black-box model, increasing training data may 
improve interpolation performance, but the model will remain unable to 
extrapolate and training data were limited to speeds under 30 mph.
The grey-box model however can extrapolate thanks to the physics-informed mixture kernel, giving it a performance advantage across all training data quantities.

Figure~\ref{fig:dan_time} shows that for a given data quantity, the grey-box model takes longer to train than the black-box model, with this difference becoming more significant as training data quantity increases.
The black-box SE kernel has three hyperparameters to learn, whereas the grey-box mixture kernel has eleven, though some of these were bounded to reduce the search space.
The optimisation problem therefore spans more dimensions so not only are there more hyperparameters to learn, but the optimisation problem itself is more complex.
The switch in the kernel mixture is also unstable, meaning the QPSO requires more generations to converge - as each generation of each particle requires an inversion of the covariance matrix, this scales the computational cost significantly, explaining the sharp increase in training time for the grey-box model.

\begin{figure}[H]
    \centering
    \subfloat[Black-Box]{%
        \includegraphics[width=0.48\linewidth]{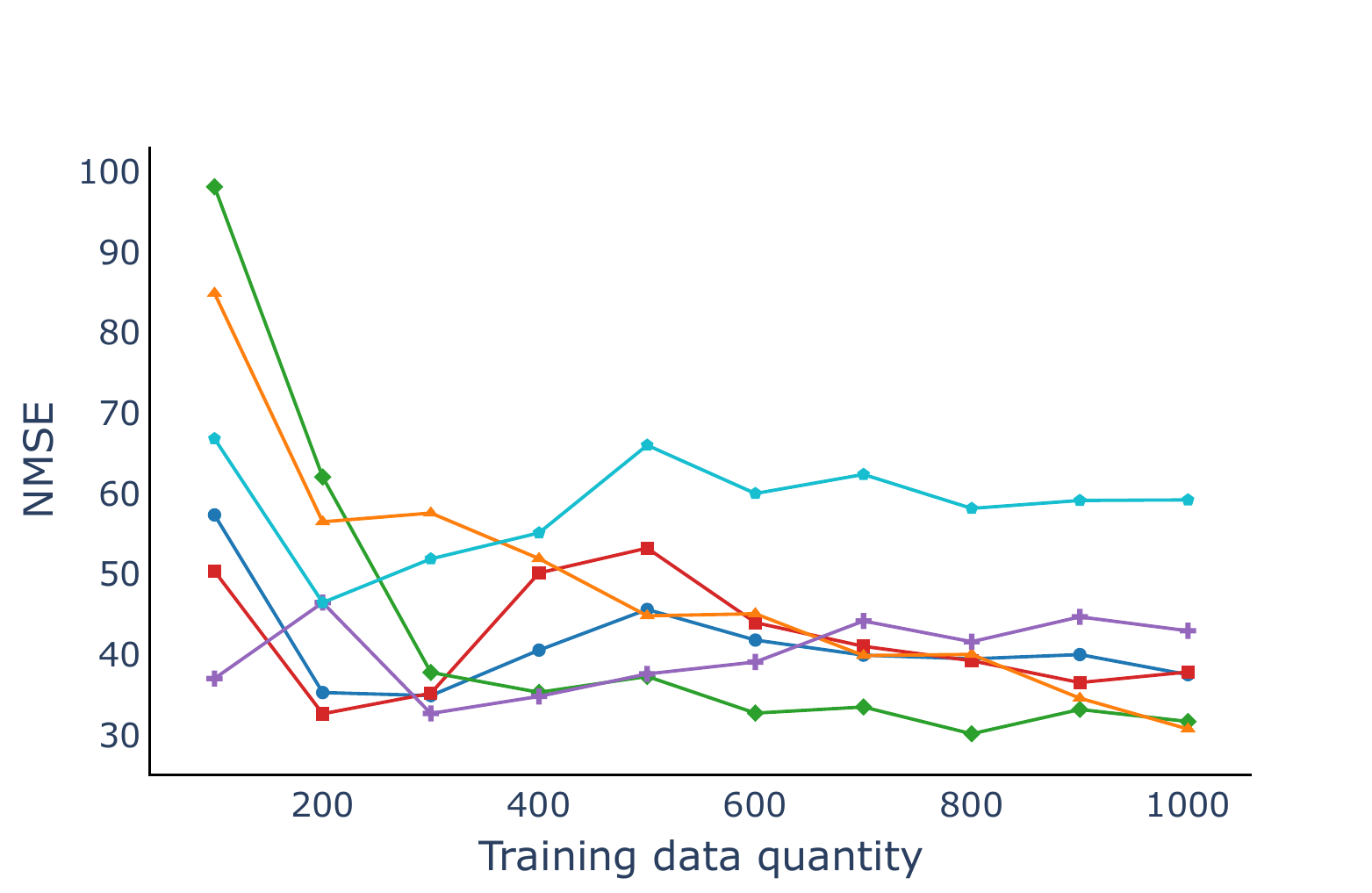}%
        \label{fig:dan_black_var}%
    }
    \hfill
    \subfloat[Grey-Box]{%
        \includegraphics[width=0.48\linewidth]{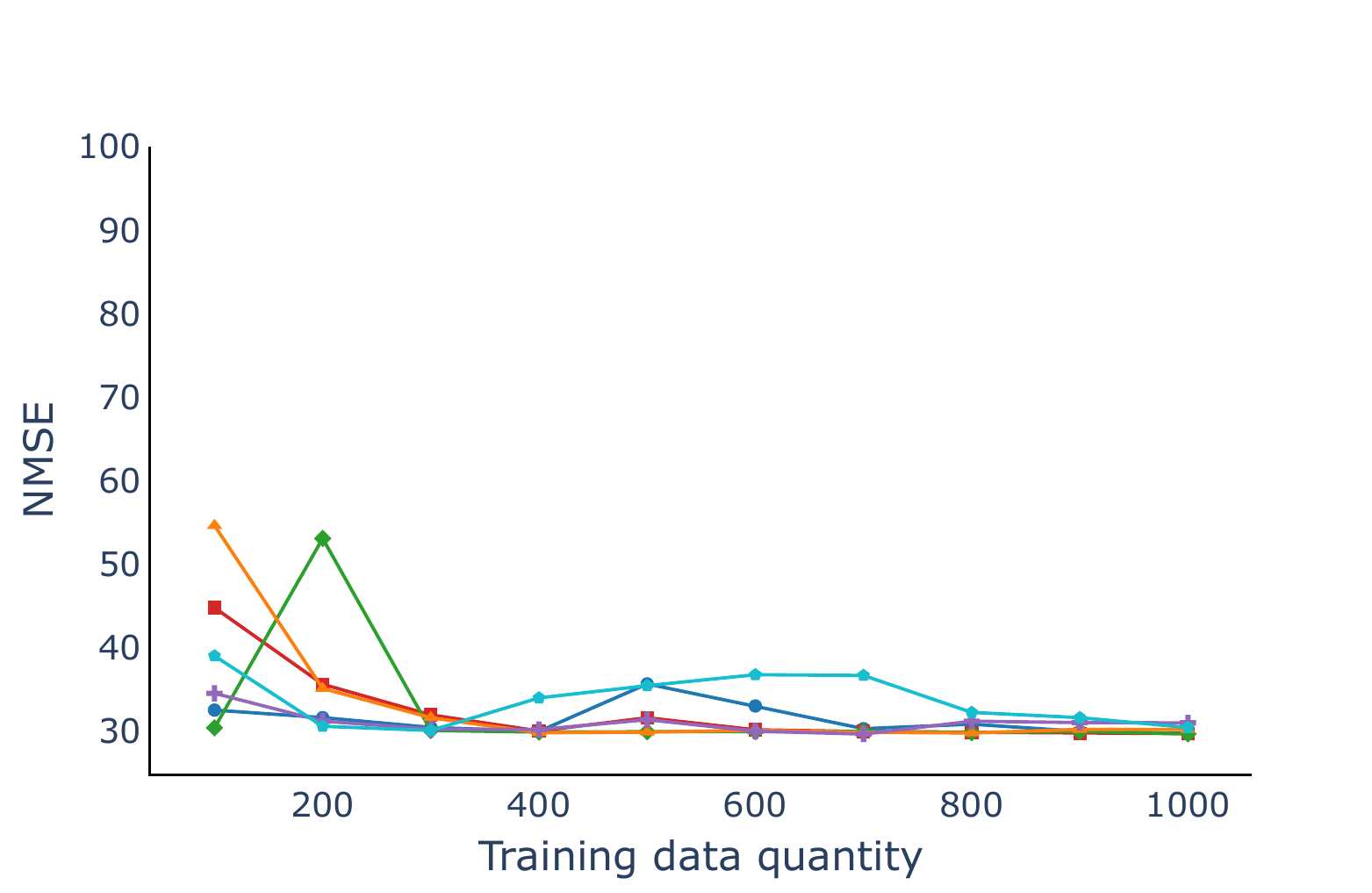}%
        \label{fig:dan_grey_var}%
    }
    \caption{The variation in NMSE for the black-box and grey-box models across six runs with different random training data selections.}
    \label{fig:dan_var}
\end{figure}

It is clear from Figure~\ref{fig:dan_time} that even with less training data, the grey-box model may have a longer training time hence higher training emissions than the black-box model.
That said, the grey-box performance with 300 data points is strong while the 
training time remains relatively low, so time/emissions savings are still 
possible. As there is no performance threshold that both models can achieve, 
with the black-box not able to match grey-box performance at any level, we 
cannot directly compare their equivalent emissions. As an additional 
consideration, Figure~\ref{fig:dan_var} shows the NMSE results for six runs 
of the black-box 
and grey-box models with different random training data selections. It shows 
that the black-box has a higher variance in performance and is more sensitive 
to training data layout, whereas the grey-box model is more robust, 
suggesting that more repeats of the black-box model may be necessary to 
ensure good model performance.

This case study highlights several important points for us to consider.
Firstly, sometimes, increasing training data only leads to marginal performance gains while significantly scaling the run time.
We must consider the necessity of these one percent gains against the penalty of doubling or tripling (or worse) our emissions from training - we will likely find them to be superfluous.
Secondly, we should consider the sensitivity of our models to training data 
layout to avoid needing to repeat training runs to achieve good performance, 
creating unnecessary emissions in the process.
Finally, where training data are few or incomplete in certain conditions,  
the grey-box models that are able to extrapolate will out-perform 
their black-box counterparts. 
Depending on our performance needs, it may be that we can implement the 
black-box model to reduce  training time and emissions, or we may require the 
grey-box advantage for a suitable model.
In any case, if possible, it would be advantageous to be aware of the 
emissions from training and consider the trade-offs before we start 
collecting training data. In this way, we can also make carbon savings from 
the collection and storage of unnecessary training data.

The final study in this paper, constrained GP modelling, is explored in the 
following section.
\section{Constrained Gaussian Processes}
\label{sec:Matty}

Constrained GPs lie towards the data-driven end of the physics-informed 
spectrum.
There are many ways to implement constraints within a GP model, each designed 
to limit the model behaviour in some way such that it aligns with the 
real-world system. This restricts the function space, as behaviours that 
violate physical constraints are ruled out, potentially decreasing the amount 
of training data required to reach a given level of accuracy.

The models used in this case study were first developed in \cite{solin2020hilbert, solin2019boundaries}, which \cite{jones2023constraining} utilised to embed physical constraints  to learn difference in time of arrival ($\Delta$T) acoustic emission (AE) maps for damage localisation.
While AE maps can be a valuable tool for SHM, and crack localisation in 
particular, they require the collection of extensive data from artificial 
sources across the whole structure.
Being able to collect data sparsely from a limited area of a structure 
alongside a machine learner allows for a more efficient and cost-effective 
approach. However irregularities in the structure, such as holes or bolts, 
disrupt the AE wave propagation, making the machine learning task challenging 
in sparse data settings. A physics-informed approach that incorporates these 
features and their effects into the model helps remedy this problem, reducing 
the need for extensive data collection and allowing for accurate damage 
localisation.

\begin{figure}[h]
    \centering
    \includegraphics[width=0.5\linewidth]{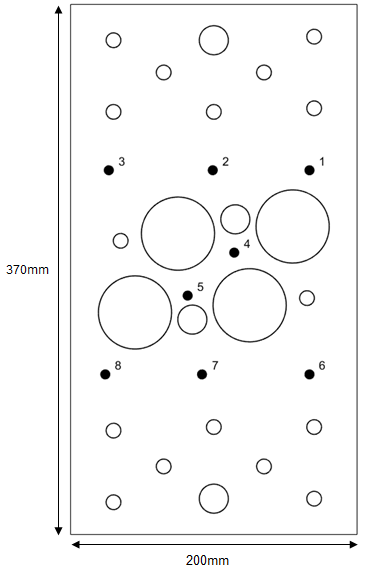}
    \caption{The plate used as a test case for the $\Delta$T AE map ML task}
    \label{fig:matty_plate}
\end{figure}

This case study used a complex plate shown in Figure \ref{fig:matty_plate}, 
originally used in \cite{hensman2010locating}.

The geometry is incorporated into the model through a sparse approximation of the kernel function, in this case using an eigendecomposition of the Laplace operator of a fixed domain \cite{solin2020hilbert}:

\begin{equation}
    k(\mathbf{x}, \mathbf{x}') \approx \sum_{j=1}^{m} S\left(\sqrt{\lambda_j}\right) \phi_j(\mathbf{x}) \phi_j(\mathbf{x}')
    \label{eq:matty_kernel_approximation}
\end{equation}

\noindent where $\phi_j$ and $\lambda_j$ are the eigenfunctions and 
eigenvalues of the Laplace operator, $S$ is the spectral density of the 
kernel, and $m$ is the number of bases used in the approximation.
To create the grey-box model, the domain was fixed to adhere to the known boundaries.
This method of approximating the kernel function reduces the computational complexity of the GP from $\mathcal{O}(n^3)$ to $\mathcal{O}(nm^2)$, where $n$ is the training data quantity and $m$ is the number of bases used in the approximation, which is fixed at 256 in this work.
Therefore, two black-box models are used for comparison: the standard GP with 
an SE kernel, and a sparse GP using the same approximation but with a wide 
domain that is away from the actual boundaries.
The sparse GP does not contain physical knowledge to represent how the sparse model form would be used in a standard black-box setting.

We look here at the case from \cite{jones2023constraining} where the grey-box 
approach has the most advantage - which is when training data available are  
from a limited section of the plate. Such a scenario is relevant in many 
engineering scenarios where we cannot access the entire structure when 
collecting data.

\subsection{Results and Discussion}

\begin{figure}[h]
    \centering
    \subfloat[NMSE]{%
        \includegraphics[width=0.48\linewidth]{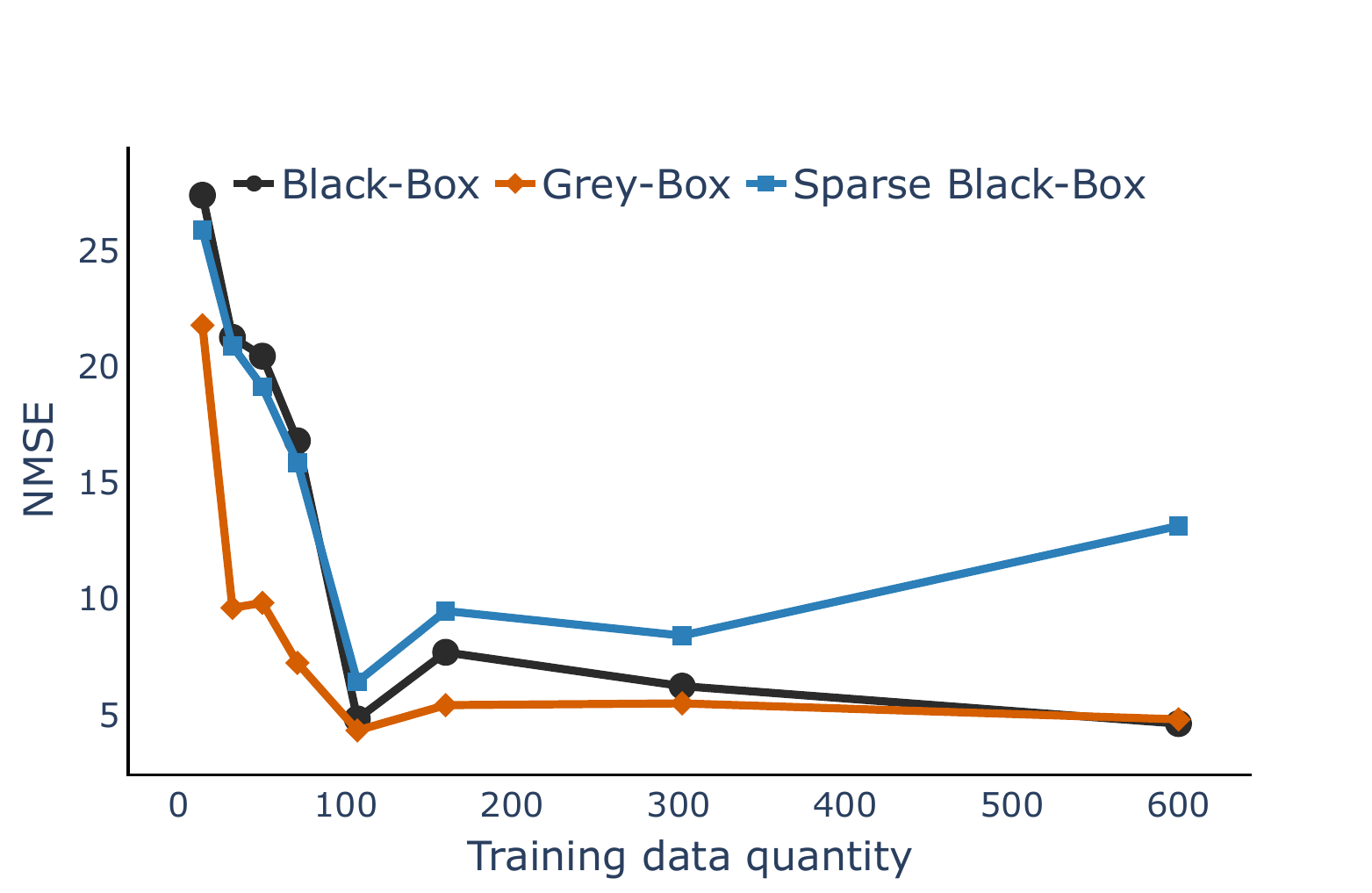}%
        \label{fig:matty_nmse}%
    }
    \hfill
    \subfloat[Training Time]{%
        \includegraphics[width=0.48\linewidth]{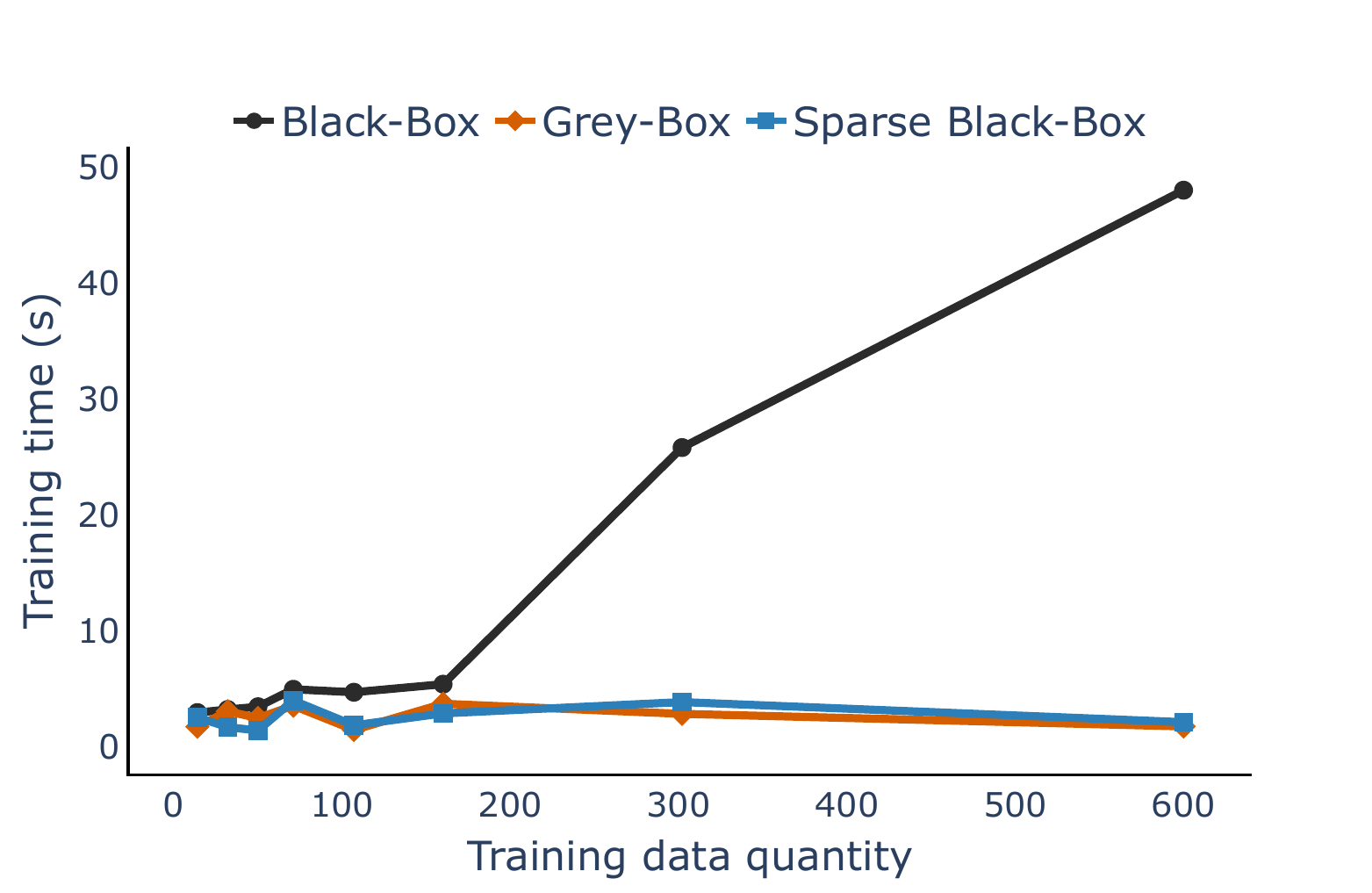}%
        \label{fig:matty_time}%
    }
    \caption{The results in terms of NMSE and training time for the 
    black-box, sparse black-box and grey-box models for the constrained GP 
    case study.}
    \label{fig:matty_results}
\end{figure}

Figure~\ref{fig:matty_results} shows the NMSE and training time results for 
the black-box, sparse black-box, and grey-box models with different training 
data quantities across a fixed domain, that is from a cross section of the 
plate. 

Figure~\ref{fig:matty_nmse} shows that the grey-box outperforms both black-box models when training data quantity is low, however the grey-box and standard black-box models converge to similar performance as the training data quantity increases, with the sparse black-box model performing the worst.
This highlights that the sparse approximation comes at a cost to performance when it does not incorporate physical knowledge, though at some thresholds, the performance of the sparse black-box model is sufficient for the task at hand.
Taking a threshold of NMSE $<$10 as an example, the grey-box achieves a 35.5\% emissions reduction compared to the standard black-box model, though the sparse black-box model sees a 58.1\% reduction in emissions (0.31 gCO$_2$e for the standard black-box, 0.20 gCO$_2$e for the grey-box and 0.13 gCO$_2$e for the sparse black-box).
Figure~\ref{fig:matty_time} shows that both sparse models have shorter 
training times than the standard black-box GP, as expected due to their 
reduced computational complexity, although this only becomes prominent at 
large training data quantities.
Therefore, we see that in regions where the grey-box performance is superior, the training time savings are insignificant, and conversely, in settings where the grey-box model sees a significant training time reduction, the performance is matched by the standard black-box model.
This shows the importance of considering training time when selecting a model - taking a training data quantity of 600 as an example, the NMSE performances are indistinguishable, yet the training time of the grey-box is 25.9 times faster, meaning the emissions from training are reduced by 96.1\% (3.11 gCO$_2$e for the standard black-box, and 0.12 gCO$_2$e for the grey-box).
As with all the case studies in this paper, the emissions themselves are small, though it is easy to imagine how these savings would upscale when considering industry applications - in this case, building localisation models for multiple, large structures would see a significant reduction in emissions when using the grey-box model.
In this case study, the constrained physics-informed GP is undoubtedly our 
green-box model, as performance is either superior or equivalent, but 
training time is always reduced, sometimes by a significant factor.

Having looked at five different case studies of PIML modelling and examining 
the impact of physics knowledge integration on emissions, the final section 
will summarise the findings and provide recommendations for future work in 
this area.
\section{Discussion and conclusions}
\label{sec:conclusions}

In this work, we examined five case studies of different PIML methods: 
residual modelling, input augmentation, hybrid models with physics-informed 
kernel selection (with and without switching), and constrained GPs. In each 
case, the physics-informed model was compared to an equivalent black-box 
model in terms of performance (normalised mean squared error) and training time to 
evaluate the impact of integrating physical knowledge on model emissions 
during training.

Across the five case studies, several common themes emerged. Firstly, as 
expected, the 
performance advantage of physics-informed models is most prominent when 
training data quantity is low. In terms of emissions, requiring less training 
data to be collected is already a saving in terms of data storage and 
transmission; the best results here also showed notable reduction in run time 
and subsequent emissions from model training. At given target errors, we saw 
gains from each of the model types explored except in the input augmentation 
case study - where adapting the inputs to the GPs and neural networks made 
little difference. 

While you can cherry pick very good results in terms of savings from the case 
studies, the results are variable if you consider different target error 
thresholds. In a number of the case studies, we demonstrated how adding 
physical knowledge can increase 
model complexity, resulting in longer training times than a black-box model 
for a given dataset size; in these cases, a reduction in training data is 
essential to achieve a net benefit in emissions. 

A third observation from the case studies is that grey-box models often 
demonstrate more consistent performance as training data increases or as the 
data layout changes, making them (in the cases shown) more reliable and 
robust than their 
black-box equivalents. In some applications, this reliability is more 
important than peak performance, making grey-box models the better choice 
even with a performance penalty. 


How does this compare to what we expect? The training time when establishing 
a regression model should broadly be a function of the complexity of the 
model, the amount of training data used, and the number of steps required to 
successfully navigate the cost surface during training. We expect that a 
physics-informed approach can help to lessen the requirement on training 
data, and indeed have demonstrated gains from this here - this is our main 
result and one that we expect to see much benefit from as we increase the 
considered training data from the relatively small sizes that we used for 
this initial work. We could particularly expect to see larger gains from this 
if the training data used requires access to the DRAM/VRAM - so long as our 
physics still holds in such cases.  

We have also seen that the increased complexity of some physics-informed 
models can counteract 
the gains from reduced data requirements, with a trade-off required between 
complexity cost and 
performance - this would of course also be true if we implemented less simple 
and more computationally expensive physics into the models. Here we saw the 
increased complexity manifest in an increased hyperparameter count in the GP 
models, lengthening the time needed during optimisation. Something that we 
have only touched briefly on here in the context of residual models, is also 
the possibility that physics could make the cost surface less navigable and 
increase runtime. Our results are positive but with a number of caveats and 
trade-offs.

When considering responsible computing, we advocate that a key first step is 
establishing the required level of model performance, something we hope is 
achievable in many engineering settings. Defining this target allows 
researchers to align their modelling strategy and dataset size with carbon 
reduction goals, avoiding the pursuit of marginal performance gains that 
often come at a heavy carbon cost. In many cases, our findings suggest that 
physics-informed models can reach target performance thresholds with lower 
emissions than their black-box counterparts, although these savings remain 
problem-dependent and currently do require case-by-case evaluation.

Because each PIML method in this study was evaluated within a single 
engineering context, these findings cannot yet be generalised across all 
domain applications or techniques. Nonetheless, this work provides a clear 
foundation for future research. Expanding these methods across broader 
engineering domains will help clarify precisely when and where specific PIML 
approaches yield green-box models. Furthermore, evaluating additional PIML 
techniques not covered here may reveal further opportunities for emissions 
reductions compared to standard black-box baselines.

\subsubsection*{Acknowledgments}
We gratefully acknowledge the support of the University of Sheffield and the EPSRC grant ROSEHIPS. Grant number EP/W005816/1.

\subsubsection*{Funding Statement}
This research was supported by grants from the EPSRC (ROSEHIPS - EP/W005816/1). MRJ gratefully acknowledges the support of the University of Sheffield through a Research Excellence
Fellowship.

\subsubsection*{Data Availability Statement}
The Toybox dataset used in this study is available at \url{https://zenodo.org/records/23667970}. The GARTEUR dataset is available at \url{https://zenodo.org/records/23668036}. Other datasets used are not able to be shared due to their commercial sensitivity. 

\subsubsection*{Ethical Standards}
The research meets all ethical guidelines, including adherence to the legal requirements of the study country.

\subsubsection*{Author Contributions}
 Conceptualization: Daisy Bradley; Elizabeth Cross. Methodology: Daisy Bradley; Nathan Hinchliffe; Daniel Pitchforth; Matthew Jones. Data curation: Daisy Bradley. Data visualisation: Daisy Bradley. Writing original draft: Daisy Bradley; Elizabeth Cross. All authors approved the final submitted draft.

\bibliographystyle{IEEEtranN}
\bibliography{references}

\end{document}